\documentclass[sigplan,10pt]{acmart}
\setcopyright{none}

\makeatletter
\def\@ACM@copyright@check@cc{}
\makeatother
\acmYear{2026}\copyrightyear{2026}
\setcopyright{cc}
\setcctype[4.0]{by}
\acmConference[EUROSYS '26]{European Conference on Computer Systems}{April 27--30, 2026}{Edinburgh, Scotland Uk}
\acmBooktitle{European Conference on Computer Systems (EUROSYS '26), April 27--30, 2026, Edinburgh, Scotland Uk}
\acmDOI{10.1145/3767295.3803572}
\acmISBN{979-8-4007-2212-7/26/04}

\usepackage{tikz}
\usepackage{amsmath}
\usepackage[compact]{titlesec}
\usepackage[labelfont=bf,font=small]{caption}[=v3.5]
\usepackage[inline]{enumitem}
\usepackage{graphicx}
\usepackage{array}
\usepackage{xcolor}
\usepackage{pifont}
\usepackage{footmisc}
\usepackage{multirow}
\usepackage{multicol}
\usepackage{subfiles}
\usepackage[normalem]{ulem}
\usepackage{booktabs}
\usepackage{makecell}

\makeatletter
\patchcmd{\maketitle}
	{\@maketitle}
	{\vspace{-1.4cm}\@maketitle\vspace{-1cm}}
	{}
	{}
\makeatother

\setlist[itemize]{topsep=2pt,partopsep=1pt,itemsep=1pt,parsep=0pt}
\setlist[enumerate]{topsep=2pt,partopsep=1pt,itemsep=0.5pt,parsep=0pt}
\definecolor{myblue}{HTML}{0E6B80}
\definecolor{mygreen}{HTML}{6B8A47}
\definecolor{myred}{HTML}{B1001C}
\definecolor{darkred}{HTML}{A62038}
\graphicspath{{./figures/}}
\usepackage[labelformat=simple]{subcaption}

\makeatletter
\renewcommand{\sectionautorefname}{\S\@gobble}
\renewcommand{\subsectionautorefname}{\S\@gobble}
\renewcommand{\subsubsectionautorefname}{\S\@gobble}
\makeatother

\newcommand{\sys}{P\footnotesize{RISM}}

\microtypecontext{spacing=nonfrench}

\begin{document}

\pagenumbering{gobble}

\date{}
\title{On-device Semantic Selection Made Low Latency and Memory Efficient with Monolithic Forwarding}

\author{Jiahao Zhou}
\affiliation{\institution{Shanghai Jiao Tong University}\country{China}}
\author{Chengliang Lin}
\affiliation{\institution{Shanghai Jiao Tong University}\country{China}}
\author{Dingji Li}
\affiliation{\institution{Huawei}\country{China}}
\author{Mingkai Dong}
\affiliation{\institution{Shanghai Jiao Tong University}\country{China}}
\author{Haibo Chen}
\affiliation{\institution{Shanghai Jiao Tong University}\country{China}}
\renewcommand{\shortauthors}{Zhou et al.}

\makeatletter
\def\@mkauthors@iii{%
  \global\setbox\mktitle@bx=\vbox{\noindent\unvbox\mktitle@bx\par\smallskip
    \centering
    \makebox[\textwidth][c]{%
      \begin{tabular}{ccccc}
      Jiahao Zhou$^{\dagger}$ & Chengliang Lin$^{\dagger}$ & Dingji Li$^{\ddagger}$ & Mingkai Dong$^{\dagger}$ & Haibo Chen$^{\dagger}$
      \end{tabular}%
    }\\[0.4em]
    $^{\dagger}$Shanghai Jiao Tong University \qquad $^{\ddagger}$Huawei\par\bigskip
  }%
}
\makeatother


\begin{abstract}

Semantic top-$K$ selection with cross-encoder rerankers underpins on-device AI services, such as retrieval-augmented generation, agent memory, and personalized recommendation.
However, its latency and memory demands dominate end-to-end budgets on edge hardware.
Revisiting the objective of top-$K$ selection, we reveal that only \emph{relative rankings} matter, not exact per-candidate scores.
We further observe \emph{sequence-level sparsity}: relative rankings progressively stabilize in intermediate layers, enabling early pruning prior to completing full inference.

Building on this insight, we propose \emph{monolithic forwarding} and develop a training-free inference system, {\sys}.
By maintaining a global view of all candidates, it reduces latency through progressive cluster pruning.
It also bounds peak memory usage by strategically overlapping I/O with computation via overlapped layer streaming and chunked execution.
We evaluate {\sys} against state-of-the-art baselines on rerankers from 0.6\,B to 8\,B parameters across Apple M2 and RTX 5070.
{\sys} consistently reduces latency by up to 89.2\% and peak memory by up to 91.3\% in microbenchmarks, without compromising precision.
Across three real-world on-device AI applications, {\sys} lowers latency by 11.6\%--51.0\% and peak memory by 18.6\%--77.8\%, demonstrating substantial improvements in efficiency and deployability.

\end{abstract}


\maketitle
\vspace{-0.4\baselineskip}
\pagestyle{plain}

\section{Introduction}%

\begin{figure}
    \centering
    \includegraphics[width=\linewidth]{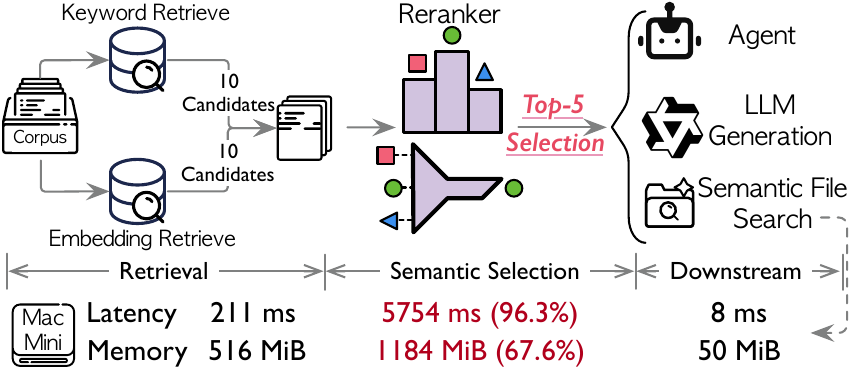}
    \vspace{-20pt}
    \caption{\textbf{Typical on-device top-$K$ selection pipeline and per-stage cost.}
    Per-stage latency and peak memory under a representative on-device semantic file search is reported.}
    \vspace{-15pt}
    \label{fig:intro_pipeline_example}
\end{figure}

Semantic selection---the process of identifying the most semantically relevant top-$K$ items from a candidate pool---serves as a core component in a wide range of on-device artificial intelligence (AI) services running on commodity PCs and laptops,
such as retrieval-augmented generation (RAG)~\cite{xia2024hybridrag,wang2024mememorag,li2025eacorag,lewis2020rag},
AI agent memory~\cite{zhang2025mobiagent,wang2025agentmemory,chhikara2025mem0,kang2025memoryos},
and personalized recommendation~\cite{google2025magiccue,darling2025magiccue}.
For example, in a typical semantic file search scenario illustrated in~\autoref{fig:intro_pipeline_example}, \emph{keyword retrieval} and \emph{embedding retrieval} select ten candidates respectively from a large corpus;
a \emph{reranker} then semantically selects the final top-$K$ items to feed downstream components such as
a large language model (LLM), a UI agent, or the user.
The precision of this top-$K$ selection directly governs downstream task quality and helps mitigate hallucinations~\cite{lewis2020rag}.

To meet the stringent precision requirements, \emph{cross-encoder rerankers}~\cite{zhang2025qwen3embedding,googlerankingapi,bge-reranker-v2-minicpm-layerwise,bge-reranker-v2-gemma} have emerged as the state-of-the-art technique for semantic selection~\cite{glass2022re2g,cohere_rerank,azure_rerank}.
Cross-encoder rerankers are transformers that process a query-candidate pair as a single, combined input and output their relevance score.
Jointly processing the query-candidate pair enables a deep, token-level analysis of their relevance through attention mechanism~\cite{attentionisallyouneed},
delivering substantial precision gains of 15\% -- 25\%~\cite{thakur2021beir, moreira2024ragreranker} over
traditional bi-encoders~\cite{weller2025embeddinglimit} that process query and candidate separately.
However, this superior precision comes at a prohibitive resource cost for on-device deployment.
For example, selecting top-5 from 20 candidates with a 0.6\,B cross-encoder reranker incurs 5,754\,ms latency and 1,184\,MB peak memory consumption on a Mac Mini.
The reranker alone contributes 96.3\% of the latency and 67.6\% of the memory footprint in the overall semantic file search pipeline, pushing the resource usage beyond typical mobile OS budgets (e.g., HarmonyOS guidance~\cite{harmonyos-memory}) and severely harming user experience.


Unfortunately, existing optimizations fall short for reducing the high latency and large memory consumption of on-device rerankers due to fundamental workload mismatch.
Decoding-centric optimizations~\cite{song2024powerinfer1,xue2024powerinfer2,kwon2023pagedattention,yanqi2024leankv} provide limited benefit for rerankers, which are prefill-only.
Methods based on \emph{token-level sparsity}~\cite{lai2025flexprefill,fu2024lazyllm,long2025sliminfer} are primarily designed for very long contexts (e.g., tens of thousands of tokens) and offer little advantage for the short, information-dense inputs typical of rerankers (e.g., up to 512 tokens).
Other approaches, such as \emph{model compression}~\cite{ma2023llmpruner} or early exit~\cite{xin2020deebert}, require retraining and task-specific tuning, complicating the deployment.
Consequently, rerankers’ high latency and resource demands severely limit the efficiency and applicability of on-device AI services.

%

In this paper, we revisit the core objective of semantic top-$K$ selection:
\emph{instead of computing absolute scores for all candidates, it suffices to identify the top-$K$ items via relative rankings.}
By inspecting intermediate scores of all candidates across layers (\autoref{fig:meta_analysis_cluster_ee}), we discover \textbf{sequence-level sparsity}:
relative rankings of candidates progressively stabilize and converge to the final rankings in intermediate layers, and the stabilized candidates can be identified via clustering.
This insight enables us to prune many candidates before full computation without sacrificing precision.
Moreover, for memory, we observe \textbf{overlap window}: the computation time of current layer is sufficient to fully overlap with loading the next layer's weights from SSD, enabling model weights offloading to SSD without latency overhead.

\begingroup\setlength{\skip\footins}{5pt plus 2pt minus 2pt}%
Driven by these insights, we propose \textbf{monolithic forwarding}, a novel inference paradigm tailored for cross-encoder rerankers.
It processes all candidates as a single, monolithic batch, maintaining a global view throughout computation, rather than splitting them into isolated batches as in conventional systems.\footnote{Vanilla systems often split inputs into multiple batches to balance computation and memory.}
This design offers opportunities to reduce inference latency via layer-wise pruning and lower memory footprint by on-demand weight loading.
\par\endgroup


Based on monolithic forwarding, we build {\sys}, a training-free system for low-latency and memory-efficient semantic selection.
{\sys} introduces \emph{progressive cluster pruning} to prune candidates without sacrificing precision.
Progressive cluster pruning enables a dynamic, three-way routing strategy: dropping hopeless candidates, accepting the winners, and continuing computation only on the remaining uncertain candidates.
For memory, {\sys} proposes \emph{overlapped layer streaming} to lower the memory footprint of layer weights, and \emph{embedding table caching} to reduce the memory consumption of the embedding layer. 
Moreover, monolithic forwarding incurs the challenge of the memory explosion of the intermediate tensor.
{\sys} solves this challenge by \emph{chunked execution}, which partitions candidates into chunks that fully utilize compute resources while keeping intermediate tensors within memory limits.

We implement and integrate all these techniques in {\sys} atop HuggingFace Transformers~\cite{hftransformers} and evaluate various reranking models (0.6\,B to 8\,B)
across platforms equipped with an RTX 5070 Laptop GPU and an Apple M2 SoC.
We evaluate {\sys} in microbenchmarks and three real-world on-device AI applications, comparing against HuggingFace Transformers~\cite{hftransformers} as the baseline.
Our microbenchmark results show that {\sys} consistently reduces latency by up to 89.2\% while lowering peak memory by up to 91.3\%.
In real-world applications, {\sys} delivers latency reductions of 11.6\%--51.0\% and peak memory savings of 18.6\%--77.8\%.
These significant improvements in both latency and memory substantially advance the practicality for on-device AI applications.

\noindent
In summary, this paper makes the following contributions:
\begin{itemize}[leftmargin=*]
\item We identify top-$K$ selection as a critical performance bottleneck in on-device AI applications and systematically analyze why existing optimizations fail for this workload.
\item We reveal two underexploited opportunities: sequence-level sparsity and I/O-computation overlap, and demonstrate the unique challenges of leveraging them for top-$K$ selection workloads.
\item We design and implement {\sys}, a training-free system with progressive cluster pruning, overlapped layer streaming, chunked execution, and embedding table caching for practical cross-encoder deployment on edge devices.
\item Extensive experiments demonstrate that {\sys} achieves up to 89.2\% latency reduction and 91.3\% memory reduction while maintaining precision across diverse benchmarks and device configurations.
\end{itemize}


\section{Background and Motivation}

\subsection{The Bi-Encoder vs. Cross-Encoder Trade-off}

Semantic top-$K$ selection, or simply top-$K$ selection, is a foundational component in modern on-device AI services, such as retrieval-augmented generation (RAG)~\cite{xia2024hybridrag,wang2024mememorag,li2025eacorag,lewis2020rag},
AI agent memory~\cite{zhang2025mobiagent,wang2025agentmemory,chhikara2025mem0,kang2025memoryos},
and personalized recommendation~\cite{google2025magiccue,darling2025magiccue}.
The reranking stage of this process is critical, as its precision directly dictates the quality of information provided to the user or consumed by an LLM.

Two primary architectures dominate the reranking landscape: \emph{bi-encoders} and \emph{cross-encoders}.
Bi-encoders encode the query and each candidate document into separate, fixed-size embeddings, and then rank candidates using similarity scores between these embeddings~\cite{karpukhin2020dense,zhao2024denseretrieval}.
While efficient, this separation imposes a fundamental precision ceiling: because query and candidate representations are generated independently, the model cannot capture fine-grained, token-level interactions critical for precise relevance estimation~\cite{weller2025embeddinglimit,thakur2021beir}.
Late-interaction models like ColBERT~\cite{khattab2020colbert} attempt to bridge this gap, but incur significant storage overhead by requiring per-token embeddings~\cite{wang2025leann} and are not generally adopted for instruction-following or reasoning-intensive tasks~\cite{weller2025embeddinglimit}.

In contrast, cross-encoders have emerged as the gold standard for reranking precision~\cite{glass2022re2g,cohere_rerank,azure_rerank}.
Existing cross-encoders can be categorized into two mainstream architectures: Encoder-only transformers with bidirectional self-attention (e.g., BERT-style)~\cite{devlin-2019,bge-reranker-v2-m3}, and Decoder-only transformers with causal self-attention (e.g., GPT-style) adapted for scoring~\cite{brown2020gpt, zhang2025qwen3embedding, bge-reranker-v2-minicpm-layerwise}.
Both architectures concatenate the query and a candidate into a single input sequence, process it through multiple transformer layers, and finally compute a scalar relevance score by applying a lightweight classifier head to the final hidden states. 
This joint encoding enables deep token-level attention between query and candidate across every transformer layer, capturing subtle semantic nuances and dependencies that bi-encoders inherently miss.
Hence, cross-encoders yield substantial and consistent gains in retrieval performance~\cite{thakur2021beir,moreira2024ragreranker}.
For on-device applications where output quality is critical, the superior precision of cross-encoders makes them indispensable.

\subsection{The Prohibitive On-Device Cost}

Despite their superior precision, cross-encoders face fundamental challenges on edge devices: their inference requires a full, compute-intensive forward pass for each query-candidate pair, a workload fundamentally misaligned with resource-constrained environments.


\textbf{Compute-bound Latency.}
A cross-encoder's latency is dominated by matrix multiplications in its transformer blocks.
For an input sequence of length $L$ (query + candidate) and a hidden dimension $D$,
the complexity of self-attention and the feed-forward network (FFN) scales as $O(L^2 \cdot D)$ and $O(L \cdot D^2)$, respectively.
Because this expensive computation must be executed independently for each of the $N$ candidates, the total latency scales linearly with $N$.
On edge devices with limited floating-point compute capabilities, this linear scaling results in delays that significantly degrade the user experience.

\textbf{Memory Footprint.}
Cross-encoders' memory demands present an equally formidable challenge, comprising two main components:
\begin{itemize}[leftmargin=*]
    \item \textit{Model Weights.} Even small sized rerankers (e.g., 0.6\,B parameters) require several gigabytes of storage.
    These weights are dominated by the stacked Transformer layers.
    For instance, in Qwen3-Reranker-0.6B, 28 Transformer layers (15\,M weights each layer) account for over 70\% of the weight memory.
    Loading these weights into the limited DRAM or VRAM of an edge device consumes a substantial portion of the system's memory budget.
    \item \textit{Intermediate Tensors.} During inference, another memory issue comes from transient intermediate tensors (e.g., for query, key, value projections, attention scores, and FFN outputs).
    The peak memory consumption from these tensors scales with the number of candidates being processed in a batch, and can easily exceed the memory budget of an application.
\end{itemize}

\subsection{Mismatch with Existing LLM Optimizations}

While recent advances in LLM inference have introduced numerous optimization techniques, a systematic analysis reveals that they are poorly aligned with the unique workload characteristics of on-device cross-encoder reranking.

\textbf{Decoding-centric Optimizations.}
A large body of research focuses on the autoregressive \emph{decoding} phase of LLMs.
Techniques such as speculative decoding and advanced KV-cache management are designed to accelerate the memory-bound, token-by-token generation process~\cite{song2024powerinfer1,xue2024powerinfer2,kwon2023pagedattention,Agrawal2024}.
Cross-encoder reranking, however, is a \emph{prefill-only} workload: it performs a single, compute-bound forward pass to produce a score for each query-candidate pair.
As there is no decoding phase, these optimizations are largely inapplicable.

\textbf{Long-context Optimizations.}
Another major line of research focuses on efficient processing of extremely long input sequences.
These methods often exploit token-level sparsity, such as sparse attention or dynamic token pruning, under the assumption that long contexts contain substantial informational redundancy~\cite{lai2025flexprefill,fu2024lazyllm}.
In contrast, reranking inputs are short and information-dense, typically consisting of a concise query and a highly relevant document chunk.
In this setting, token sparsity offers negligible benefit and may even compromise the fine-grained relevance judgments that cross-encoders are designed to capture.

\textbf{Training-based Compression.}
Techniques such as model pruning, quantization-aware training (QAT), and early exit can reduce inference costs, but they require expensive retraining or fine-tuning to preserve model precision~\cite{ma2023llmpruner,sanh2020distilbert,xin2020deebert,chen2024efficientqat}.
The associated cost and complexity pose a substantial barrier to rapid and reliable on-device deployment.

\textbf{Post-training Quantization.}
While 4-bit weight quantization is a common baseline optimization~\cite{frantar2023gptq,lin2024awq}, achieving practical speedups through more aggressive sub-4-bit quantization on prefill workloads remains an open challenge.
Beyond precision degradation, most edge devices also lack the specialized hardware and kernel support required for high-throughput sub-4-bit matrix multiplication, limiting real-world performance gains~\cite{bitnet_a4_8,bitnet_b1_58}.

Taken together, these analyses reveal a clear research gap: a training-free, workload-specific optimization paradigm is required to make high-precision cross-encoder reranking feasible on edge devices.


\section{{\sys} Overview}
In this section, we introduce {\sys}, a training-free inference system for cross-encoder rerankers on edge devices.

\begin{figure}[bt]
    \begin{subfigure}{\linewidth}
        \centering
        \includegraphics[width=\linewidth]{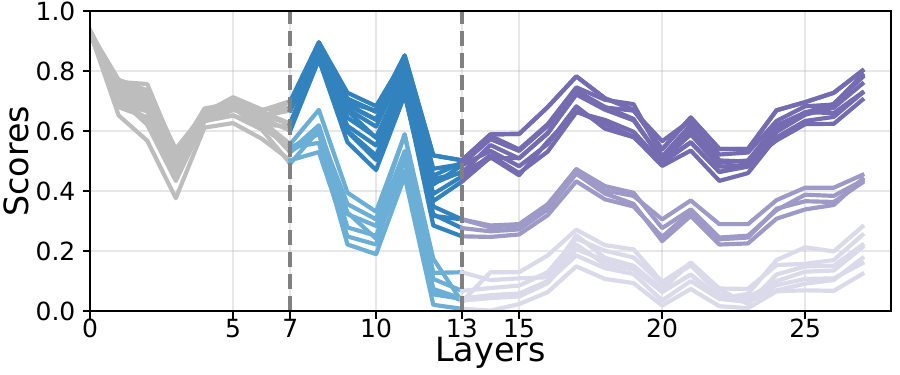}
        \caption{\textbf{Score evolution across layers reveals sequence-level sparsity.}
        Each line presents the score evolution of a candidate (20 in total) on BGE-Minicpm model.
        Candidates progressively diverge into statistically distinct clusters as they pass through layers,
        and relative rankings progressively stabilize and converge to the final rankings in intermediate layers.}
        \label{fig:meta_analysis_example}
    \end{subfigure}
    \begin{subfigure}{\linewidth}
        \centering
        \includegraphics[width=\linewidth]{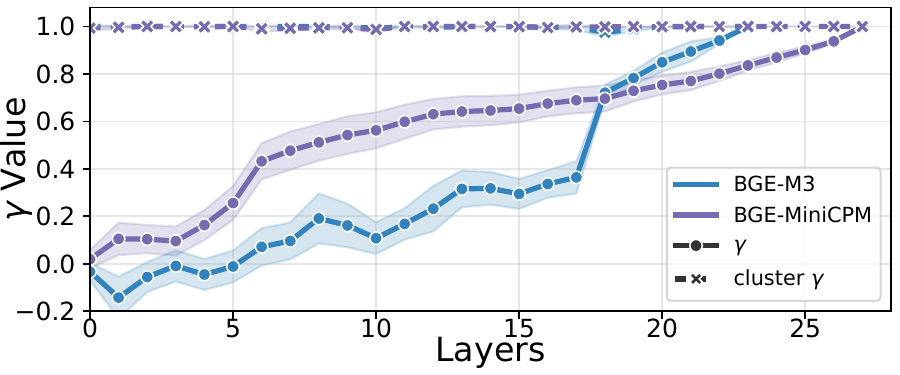}
        \caption{\textbf{Generality of sequence-level sparsity.} On 18 datasets and all mainstream model architectures, Goodman and Kruskal’s $\gamma$ rises as layers deepen and cluster $\gamma$ is consistently close to 1.0 across layers, indicating relative rankings stabilize early and the stabilized candidates can be identified by clustering and safely pruned without sacrificing precision.}
        \label{fig:meta_analysis_generality}
    \end{subfigure}
    \vspace{-20pt}
    \caption{\textbf{Key observation of sequence-level sparsity.}}
    \vspace{-20pt}
    \label{fig:meta_analysis_cluster_ee}
\end{figure}

\begin{figure*}[bt]
    \centering
    \includegraphics[width=0.88\linewidth]{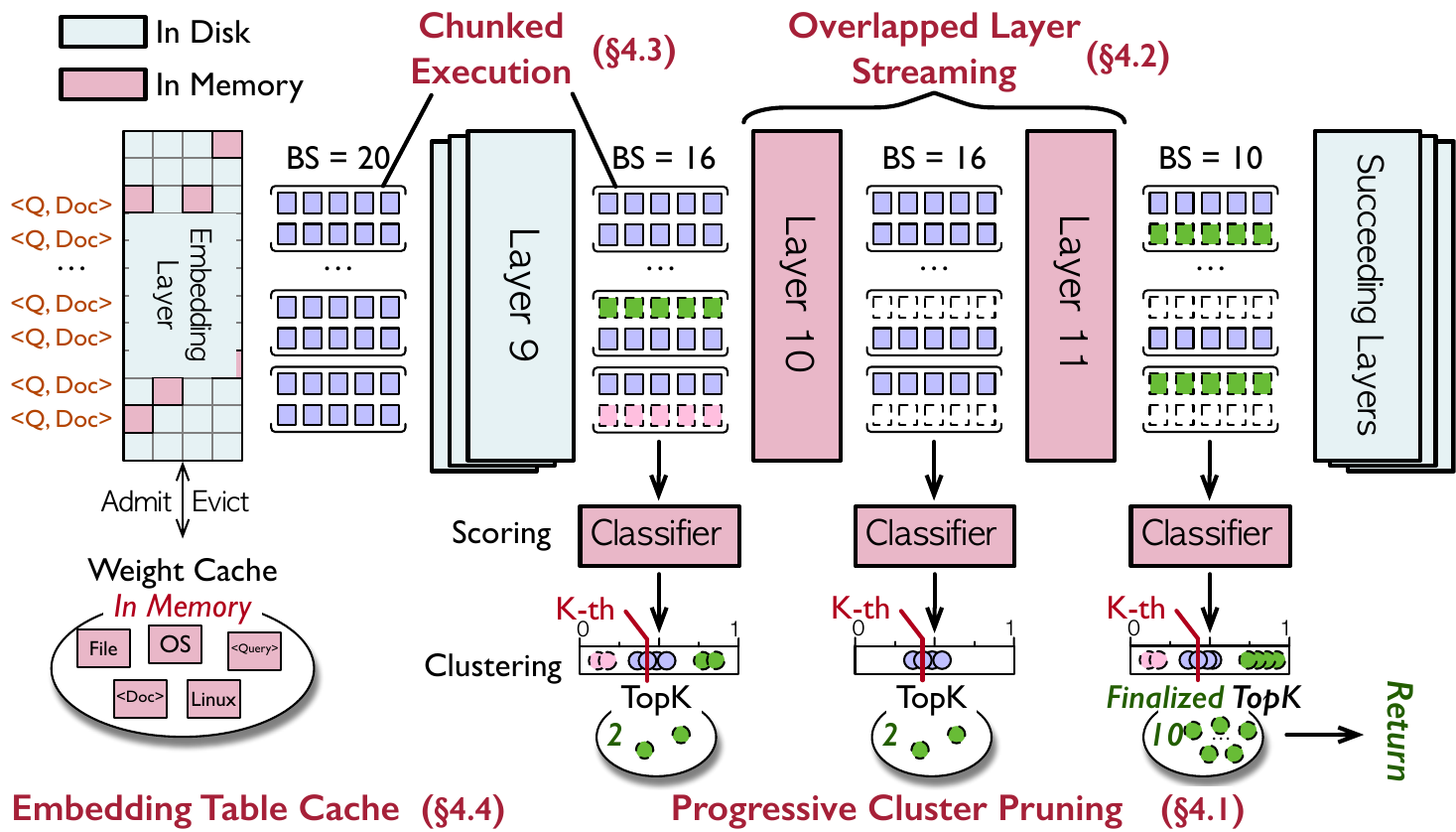}
    \caption{\textbf{{\sys} overview with a working example.}}
    \label{fig:working-example}
\end{figure*}
\subsection{Key Insight: Sequence-level Sparsity}
Our approach stems from reconsidering the fundamental objective of top-$K$ selection:
\emph{identifying relative rankings among candidates, rather than computing precise absolute scores}.
This insight motivates us to examine how candidate rankings evolve across transformer layers from a global perspective.

In~\autoref{fig:meta_analysis_example}, we compute candidate scores at each layer using the model's original classifier and show score evolution across layers.
We observe that candidate scores progressively diverge into statistically distinct clusters as they pass through layers.
While the relative rankings of candidates within the same cluster remain in flux (lines of the same color), the relative rankings \emph{between} different clusters stabilize early and converge to the final rankings (lines of different colors).
This motivates our insight of \textbf{sequence-level sparsity}:
relative rankings of candidates progressively stabilize and converge to the final rankings in intermediate layers, and the stabilized candidates can be identified by clustering and safely pruned without sacrificing precision.

We validate the generality of sequence-level sparsity across 18 datasets (detailed in \autoref{sec:experiment-setup}) and all mainstream model architectures (encoder- and decoder-only).
We quantify ranking convergence using Goodman and Kruskal's $\gamma$~\cite{wiki2022gamma}, computed over candidate pairs by counting those whose relative order is preserved between an intermediate and the final layer (concordant, $N_c$) versus reversed (discordant, $N_d$):
$
\gamma = \frac{N_c - N_d}{N_c + N_d}.
$
To directly measure the stability of inter-cluster rankings, we further define \emph{cluster $\gamma$}, which restricts the computation to only candidate pairs from different clusters.
\autoref{fig:meta_analysis_generality} shows that the standard $\gamma$ consistently rises as layers deepen (lines with circular markers), confirming that relative rankings progressively converge.
More importantly, the cluster $\gamma$ remains consistently close to 1.0 across all layers (lines with cross markers), strongly validating that stabilized candidates can be identified by clustering and safely pruned without sacrificing precision.

We attribute the sequence-level sparsity to model's coarse-to-fine understanding~\cite{jawahar2019bert, sajjad2022analyzing}.
Early layers capture broad semantic and separate candidates with clear relevance differences into distinct clusters; as layers deepen, these coarse clusters progressively split into finer ones to resolve subtle distinctions among highly similar candidates.
As a result, inter-cluster rankings stabilize early, enabling us to safely prune while preserving precision.

\subsection{Key Insight: Overlap Window}
The prefill-only nature of reranking creates an opportunity for aggressive memory optimization.
Unlike autoregressive generation that processes tokens iteratively, reranking performs a single forward pass over all tokens of a candidate, yielding high arithmetic intensity per layer.
Meanwhile, although edge devices offer limited compute throughput, SSDs sustain high bandwidth.
The combination of compute-heavy layers and fast storage opens an \textbf{overlap window}: the computation time of the current layer is sufficient to fully overlap with loading the next layer's weights from SSD, enabling model weights offloading to SSDs without latency overhead.

\subsection{Monolithic Forwarding}
Motivated by these insights, we propose \textbf{monolithic forwarding},
a novel execution paradigm for cross-encoder rerankers.
Instead of processing candidates in isolated batches as conventional systems do,
monolithic forwarding consolidates all candidates into a single, unified batch 
that progresses through layers together.

This paradigm unlocks two critical opportunities.
First, maintaining a global view of all candidates throughout execution 
enables dynamic pruning based on relative rankings at each layer.
We thus can eliminate candidates that have no chance of reaching the top-$K$,
reducing computation as the forward pass proceeds.
Second, the large, consolidated batch creates substantial computation windows at each layer,
sufficient to completely overlap I/O latency of loading the next layer's weights from disk.
This allows us to keep only two layers in memory, reducing the memory footprint of model weights.

\subsection{System Overview of {\sys}}

{\sys} realizes the main idea of monolithic forwarding through several complementary techniques as illustrated in \autoref{fig:working-example}.

Before stepping forward to layer $i+1$, {\sys} leverages \textbf{progressive cluster pruning} (\autoref{sec:progress-cluster-pruning}) to prune candidates.
It first applies a clustering-based analysis to layer $i$'s output scores using statistical properties of inter-cluster separation rather than absolute score gaps,
to determine whether a stable relative ranking has emerged.
After determining whether a stable relative ranking has emerged, it routes candidates to \emph{selected}, \emph{dropped}, or \emph{deferred}.

Once layer $i+1$ starts to execute, {\sys} utilizes \textbf{overlapped layer streaming} (\autoref{sec:dual-layer}) to minimize the memory footprint of model weights~\cite{hfaccelerate}.
It immediately releases the weights of layer $i$ from memory and starts prefetching the weights of layer $i+2$ from storage.

During the execution of layer $i+1$, consolidating all candidates into a single batch inflates intermediate tensor sizes, significantly increasing peak memory consumption.
To address this challenge, {\sys} adopts \textbf{chunked execution} (\autoref{sec:chunked-execution}).
It partitions the monolithic batch into smaller chunks and executes them sequentially within each layer,
significantly reducing peak memory usage by only keeping one chunk's intermediate tensors in memory.
Simultaneously, it provides sufficient computation window to overlap I/O of loading layer $i+2$'s weights.
For extreme memory constraints, {\sys} further supports dynamic offloading of hidden states.

Additionally, {\sys} complements these techniques with \textbf{embedding table caching} (\autoref{sec:embedding-table-caching}) that exploits token distribution sparsity before layer $0$,
which substantially lowers the memory consumption of the embedding layer (\autoref{fig:ablation_memory_latency}).


\section{Detailed Design}
\begin{figure}[bt]
    \centering
    \includegraphics[width=\linewidth]{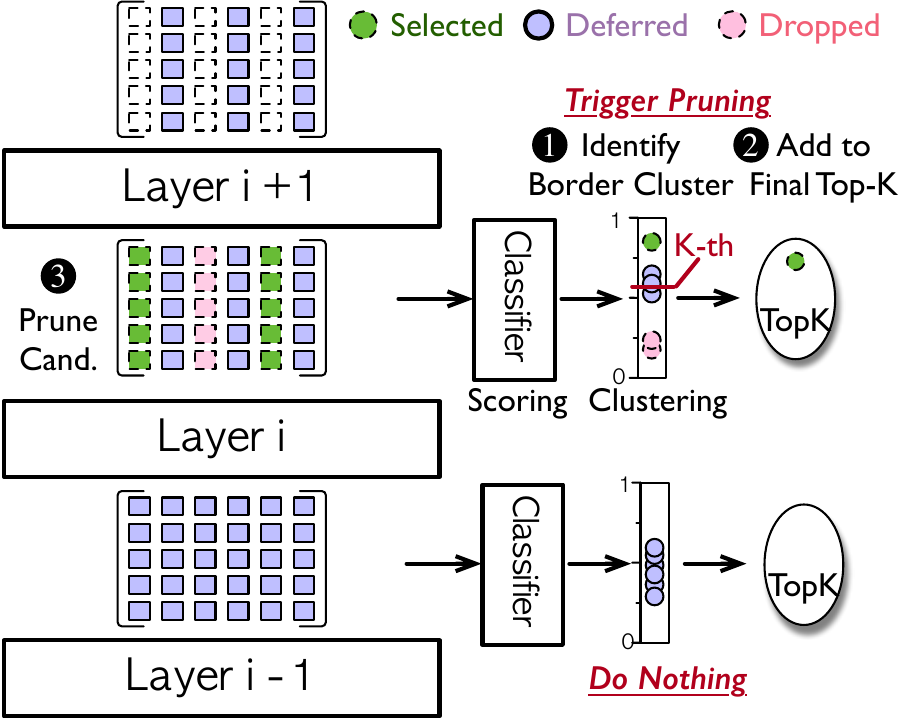}
    \vspace{-15pt}
    \caption{\textbf{Progressive cluster pruning.} Once the score dispersion exceeds the threshold during the layer progressing, we partition candidates into selected, deferred, and dropped clusters. Only the deferred candidates are retained and others are pruned.}
    \vspace{-15pt}
    \label{fig:design-cluster-ee}
\end{figure}
\subsection{Progressive Cluster Pruning}
\label{sec:progress-cluster-pruning}
To reduce the latency while preserving precision, we propose progressive cluster pruning based on sequence-level sparsity.
In each layer, we calculate the candidate scores, which fall into clusters.
Based on whether a cluster is in or out of the top-$K$ set, we decide the fate of the candidates inside: either continue computation or prune them, greatly reducing computation and thus lowering latency.

\autoref{fig:design-cluster-ee} illustrates the progressive cluster pruning.
To realize this insight, we first identify if the stable relative ranking occurs in a layer. 
In details, we employ the model's classifier to calculate the current scores of candidates.
We compute the coefficient of variation (CV)~\cite{wiki2025cv} of scores to quantify their dispersion: $CV = |\frac{std(scores)}{mean(scores)}|$.
For layer $i$-1, the CV does not exceed a predefined threshold (referred to as the dispersion threshold), we consider a stable relative ranking has not yet emerged. 
We do nothing and continue the forwarding of layer $i$. For layer $i$, the CV exceeds the dispersion threshold, we consider a stable relative ranking emerges and triggers the core clustering and pruning logic.

At the beginning, we perform K-Means~\cite{hartigan1979kmeans} on CPU to partition candidates into clusters with negligible latency overhead ($\sim$1\,ms).
The pruning logic operates at a cluster granularity. 
The process pivots on identifying the boundary cluster, which contains the K-th ranked candidate.
This boundary acts as a clear demarcation line, allowing us to classify all candidates into three groups: selected, deferred, and dropped.
Selected candidates are those in clusters with scores higher than the boundary cluster's.
The selected candidates are safely included in the final top-$K$ set and their computation ceases.
Conversely, dropped candidates are those in lower-scoring clusters.
The dropped candidates are pruned, as they have no chance of reaching the top-$K$.
Consequently, only a small subset of candidates within the boundary cluster are deferred for continued processing in subsequent layers.
Progressive cluster pruning allows the model to cease computation for the vast majority of candidates when stable relative rankings emerge, and the forward pass terminates completely if the number of deferred candidates is equal to the remaining top-$K$ slots to be filled.

It's worth mentioning that the dispersion threshold provides direct and intuitive control over the precision-latency trade-off.
A lower threshold enables more aggressive pruning, maximizing performance at a potential precision cost, whereas a higher value preserves precision by being more conservative.
Crucially, our system allows users to either manually tune this threshold or simply specify a minimum precision target. 
In the latter mode, our system automatically calibrates the threshold to the lowest possible value that meets the constraint, thereby maximizing performance under the given requirement.
In detail, we sample requests at a frequency and log their top-$K$ results.
When the device is idle, we re-execute full inference (without pruning) to obtain the ground truth.
We then compute the precision of the sampled requests against the ground truth.
If the precision falls below the target precision, we raise the dispersion threshold for precision; otherwise, we lower it for performance.

In summary, progressive cluster pruning effectively reduces the latency while maintaining precision, and we provide the system ability to navigate the precision-latency spectrum.

\begin{figure}[bt]
    \centering
    \includegraphics[width=.8\linewidth]{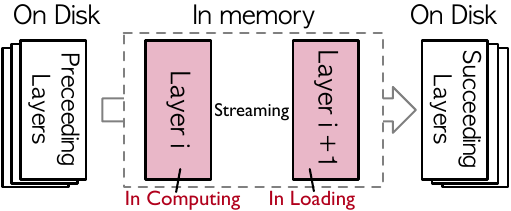}
    \vspace{-10pt}
    \caption{\textbf{Overlapped layer streaming.} Throughout the inference, we only reserve two memory buffers to stream model weights continuously. When layer $i$ resident in one buffer is computing, the next layer $i{+}1$ is prefetched from storage into the other buffer to perfectly overlap I/O. Once layer $i$ finishes computing, its buffer is released and recycled to prefetch layer $i{+}2$, sustaining a seamless streaming of weights that completely hides the load latency.}
    \label{fig:design-pipeline-offloading}
    \vspace{-10pt}
\end{figure}
\subsection{Overlapped Layer Streaming}
\label{sec:dual-layer}
To reduce the memory footprint of model weights and overlap the I/O latency of weight loading, we introduce the overlapped layer streaming technique.
This technique maintains at most two layers' weights in memory --- the current layer in computing and the next layer in prefetching --- and overlaps weight prefetching with computation to hide the I/O latency.

The core technique are depicted in~\autoref{fig:design-pipeline-offloading}.
While forwarding the layer $i$, we concurrently prefetch the weights for layer $i+1$ from disk into a dedicated memory buffer.
Thanks to the key idea of monolithic forwarding, upon the completion of layer $i$'s computation, the layer $i+1$ has already been loaded into memory and ready for forwarding.
At this time, the weights of layer $i$ are obsolete and immediately released from its memory buffer.
This vacated buffer is then recycled for the prefetching of layer $i+2$.
Now, we stream the model weights: we compute layer $i+1$ while prefetching layer $i+2$.
Throughout, only two pre-allocated memory buffers are needed to hold weights in streaming manner, significantly reducing the memory footprint.
Therefore, we perfectly overlap the computation of the current layer with the I/O of the next layer and hence incur no latency penalty.

In summary, the overlapped layer streaming minimizes the memory footprint of model weights with no latency penalty.

\subsection{Chunked Execution}
\label{sec:chunked-execution}
Monolithic forwarding incurs the challenge of the memory explosion of the intermediate tensor.
Consolidating all candidates inflates intermediate tensor sizes proportionally.
For 60 candidates with 512-token sequences on a 0.6\,B model,
intermediate tensors per layer increase peak memory by 473\,MB (see \autoref{fig:ablation_memory_latency}),
which can cause out-of-memory issues on devices with strict memory constraints.
This tension between the batch size needed for I/O overlap 
and the memory constraints of edge platforms must be carefully balanced.

To solve this challenge, we propose chunked execution. 
Our key observation is that I/O overlapping only depends on the total computation time of a layer, not necessitating executing all computations simultaneously.
Thus, we split the monolithic batch into chunks, maintaining I/O overlap while keeping only one chunk's intermediate tensors in memory, reducing peak memory usage.

\begin{figure}[tb]
    \centering
    \includegraphics[width=0.8\linewidth]{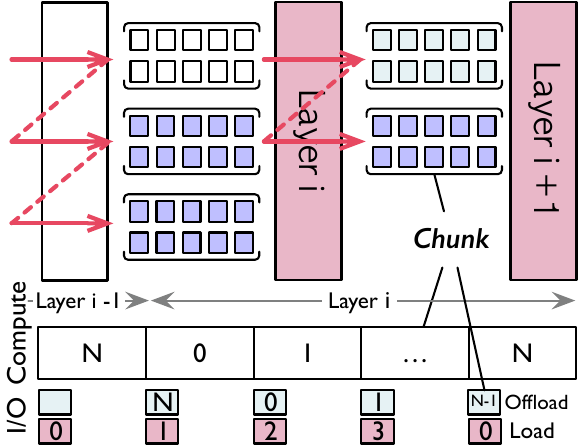}
    \vspace{-10pt}
    \caption{\textbf{Chunked execution.} For solving the memory explosion, we partition the monolithic batch into smaller chunks and execute their forwarding sequentially. To enable scalability with massive candidates, we support dynamic offloading of hidden states.}
    \vspace{-10pt}
    \label{fig:design-chunked-execution}
\end{figure}

As shown in~\autoref{fig:design-chunked-execution}, the layer's forward pass is executed sequentially on these chunks. 
Through this approach, we only need to allocate memory for the intermediate tensors of a single chunk and the hidden states of all chunks.
This maintains minimal memory footprint compared to processing the monolithic batch at once.
Notably, to fully exploit hardware computational capabilities, the chunk size has a lower bound.
We dynamically determine the optimal chunk size considering device compute capability, model size, and input sequence length.

While chunked execution effectively manages intermediate tensors, the aggregated hidden states can become a memory bottleneck when candidate number scales.
To address this, we support dynamic offloading of the hidden states.
In the lower part of~\autoref{fig:design-chunked-execution}, while computing the current chunk, we concurrently offload the completed hidden states from the previous chunk and prefetch the hidden states needed for the next chunk.
This approach ensures at most three chunks reside in memory: one being computed, one being offloading, and one being prefetched.
By bounding the memory footprint required for hidden states, we enable the scalability with massive candidates.

\subsection{Embedding Table Caching}
\label{sec:embedding-table-caching}
After the overlapped layer streaming and chunked execution significantly reduce the memory footprint of transformer layers, the embedding layer becomes the new dominant memory bottleneck.
For further optimization, we propose embedding table cache that exploits the sparsity of the embedding layer.

Consider our optimized \texttt{Qwen3-Reranker-0.6B}, the active layers consume only 60\,MB while the embedding table requires 296\,MB, accounting for over 83\% of the total memory footprint.
To address this, we observed that the activation of the embedding layer weights is highly sparse.
For the same 0.6\,B model with a vocabulary of 151,669 tokens, a typical reranking task involving 20 documents with 512 sequence length accesses 10,240 unique tokens at most, merely 6.75\% of the vocabulary.
It indicates the activation of embedding layer is highly sparse and inspires our embedding table caching.

\begin{figure}[t]
    \centering
    \includegraphics[width=\linewidth]{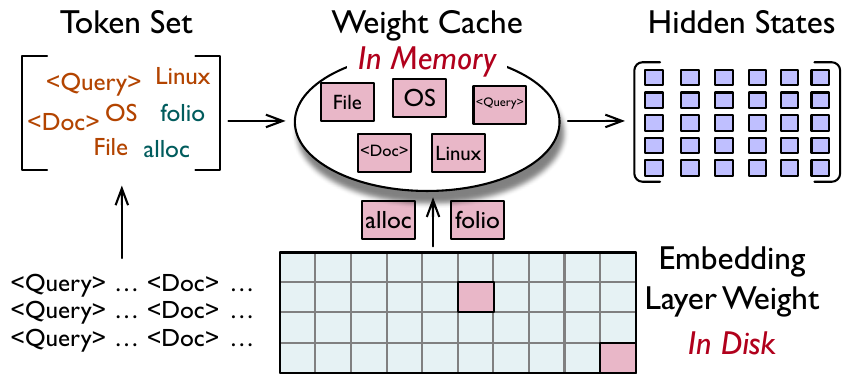}
    \vspace{-20pt}
    \caption{\textbf{Embedding table caching.} Based on the high sparsity of the embedding layer, we employ a small LRU cache to store a subset of the weights of the embedding layer. It significantly reduces the memory footprint with only a negligible latency.}
    \vspace{-15pt}
    \label{fig:design-ondemand-embedding}
\end{figure}

\autoref{fig:design-ondemand-embedding} illustrates our design. 
The core component is a small LRU cache residing in memory, which stores a subset of the embedding weights.
During inference, we first collect the set of unique input tokens and lookup the cache for the activated weights of embedding layers.
For any cache miss, the system triggers a synchronous read operation to fetch the missing weights from the disk and load them into cache.
In practice, we set the cache size to only 10\% of the vocabulary size, which significantly reduces memory consumption while maintaining high hit rates due to the skewed token distribution in natural language~\cite{zipf1949human}.
Besides, the cache miss incurs negligible latency due to the small data volume of the sparse activated weights and the effective LRU cache (see the ablation study in~\autoref{sec:abalation}).

In summary, embedding table caching technique employs a small, in-memory LRU cache to hold only the active weights of embedding table, drastically reducing the memory footprint of the full embedding table.

\subsection{A Working Example}
\autoref{fig:working-example} illustrates our system's end-to-end workflow using a typical re-ranking task as an example: identifying the top-10 most relevant documents from 20 candidates.
Before forwarding, we consolidate all candidates into one monolithic batch.
The forwarding begins at the embedding layer, where input tokens are converted into hidden states. 
To manage memory, we maintain a fixed-size cache for embedding weights. 
We identify unique input tokens not present in the cache, load them synchronously from disk, and evict existing entries using an LRU policy if the cache capacity is exceeded.

Subsequently, the hidden states are partitioned into chunks and processed through Transformer layers sequentially, where our system keeps an overlapped layer streaming across Transformer layers.
We partition one monolithic batch with 20 candidates into ten chunks with two candidates and do the forwarding sequentially for each layer.
During the forwarding of one layer, a dedicated I/O process begins prefetching the weights of the next layer from the disk in parallel.
Upon the completion of the forward pass of a layer, its weights are immediately released from the memory.
Assuming we have just completed forwarding of layer 9, we first immediately deallocate its weights from the memory and the weights of layer 10 has already been loaded into memory for the subsequent forwarding.

Critically, before forwarding each Transformer layer, we perform a cluster-based pruning check to prune candidates whose final ranks are already evident.
Continuing our example, before executing layer 10, we compute provisional scores for all 20 active candidates with the classifier layer.
Their CV exceeds a predefined threshold, indicating significant score divergence and triggering a K-Means clustering to partition candidates into multiple clusters.
To classify the clusters, we first identify the pivotal cluster containing the K-th ranked candidate (the 10th in this case).
The clusters with a mean score higher than that of the pivotal cluster are selected clusters.
Two candidates within these clusters are selected into the final top-10 and exit subsequent forwarding.
Conversely, those clusters with a lower mean score are the dropped clusters and all two candidates are dropped.
In this instance, the process identifies two selected and two dropped candidates.
Consequently, we prune these four and proceed to layer 10 with only the remaining 16 deferred candidates.

After layer 10 completes, the CV of these 16 candidates falls below the threshold, so no pruning occurs, and all proceed to layer 11.
After executing layer 11, however, the score CV once again surpasses the threshold, triggering another clustering. 
This time, a terminal condition is met: the number of candidates in the final deferred cluster, when added to the number of already included candidates, precisely equals the target $K = 10$. The system therefore terminates inference immediately, returning the combined set as the final result.


\section{Implementation}
\label{sec:impl}
We implement {\sys} in \textasciitilde5\,k lines of Python and \textasciitilde1.7\,k lines of C.
Our implementation builds on HuggingFace Transformers v4.52.4~\cite{hftransformers} and HuggingFace Accelerate v1.6.0~\cite{hfaccelerate}.

To achieve high-performance, parallel I/O, we incorporate the following implementation optimizations.
First, to bypass Python's Global Interpreter Lock (GIL)~\cite{pythongil, wiki2025gil} and thus parallelize computation and I/O operations, we spawn a computation process and an I/O process.
The two processes communicate with low latency via a shared memory buffer managed by Pytorch Multiprocessing~\cite{pytorch2025multiprocessing}.
Second, to saturate disk bandwidth, the I/O process leverages Libuv~\cite{libuv} to perform high-throughput asynchronous disk I/O.
Finally, we enable the CUDA Multi-Process Service (MPS)~\cite{nvidiamps} to facilitate efficient GPU sharing between two processes, minimizing the context-switching overhead.

\section{Evaluation}
The goal of evaluation is to answer three key questions:
\begin{itemize}[leftmargin=*]
    \item \textbf{Latency Reduction.} Can {\sys} significantly reduce latency while preserving model precision?
    \item \textbf{Memory Efficiency.} Can {\sys} substantially lower memory footprint without introducing latency overhead?
    \item \textbf{Ablation Study.} What is the individual contribution of each of the four proposed techniques to the overall performance improvement?
\end{itemize}

\subsection{Experiment Setup}
\label{sec:experiment-setup}
\paragraph{Hardware configuration.}
Experiments are conducted on two distinct platforms, representing both unified and non-unified memory architectures:
\begin{itemize}[leftmargin=*]
    \item \textbf{NVIDIA Platform.} A laptop with an Intel(R) Ultra9-275HX processor, 32\,GiB memory, NVIDIA RTX 5070 Laptop GPU with 8\,GiB memory, and 1\,TiB PCIe 4.0 SSD.
    \item \textbf{Apple Platform.} A Mac Mini with an Apple M2 SoC, 16\,GiB unified memory, and 256\,GiB PCIe 4.0 SSD.
\end{itemize}

\paragraph{Compared systems.}
We compare {\sys} with these baselines in evaluation.
\begin{itemize}[leftmargin=*]
    \item \textbf{{\sys}.} Our proposed system, which integrates all techniques: progressive cluster pruning, overlapped layer streaming, chunked execution, and embedding table caching.
    \item \textbf{HF.} The vanilla HuggingFace Transformers~\cite{hftransformers} with Pytorch backend. This baseline represents the standard, in-memory inference performance. We chose it over vLLM and SGLang due to its broader compatibility with diverse edge devices~\cite{vllm2025edge, sglang2025edge}.
    \item \textbf{HF Offload.} The vanilla HuggingFace Transformers~\cite{hftransformers} with the HuggingFace Accelerate~\cite{hfaccelerate} library's disk offloading feature. All transformer layers are offloaded to disk and loaded right before execution.
    \item \textbf{HF Quant.} The state-of-the-art quantization method. We quantize the model in W4A16 with GPTQ~\cite{frantar2023gptq}.
    \item \textbf{{\sys} Quant.} The state-of-the-art quantization method. We integrate our techniques with quantization techniques, which are orthogonal.
\end{itemize}

\begin{table}[t]\footnotesize
    \centering
    \caption{\textbf{The evaluated models.}}
    \vspace{-10pt}
    \begin{tabular}{l|l|l}
        \hline
        Name                  & Model Size & Architecture \\ \hline
        Qwen3-Reranker-0.6B   & 0.6\,B       & Decoder-only  \\
        Qwen3-Reranker-4B     & 4\,B         & Decoder-only  \\
        Qwen3-Reranker-8B     & 8\,B         & Decoder-only  \\
        Bge-Reranker-v2-MiniCPM & 2\,B       & Decoder-only  \\
        Bge-Reranker-v2-M3    & 0.6\,B         & Encoder-only  \\ \hline
    \end{tabular}
    \vspace{-15pt}
    \label{tab:evaluated-models}
\end{table}

\paragraph{Models.}
As presented in \autoref{tab:evaluated-models}, we evaluated a wide range of state-of-the-art models.
These models vary in size from 0.6\,B to 8\,B and feature diverse architectures, from encoder-only (e.g., Bge-Reranker-v2-M3) to decoder-only (e.g., the Qwen3-Reranker series).

\paragraph{Workloads.}
We evaluate the compared systems in both microbenchmarks and real-world evaluations.
In the microbenchmarks, we evaluate compared systems on 18 datasets: 15 datasets in BEIR benchmark~\cite{thakur2021beir, beir2025github}, LoTTE dataset~\cite{santhanam2022colbertv2}, Wikipedia dataset~\cite{ellamind2023wiki} and CodeRAG dataset~\cite{wang2025coderag}.
In real-world evaluations, we evaluate our system in three real-world scenarios, including RAG, Agent Memory, and LLM long context selection.
The detailed descriptions are shown in~\autoref{tab:realworld-workload-desc}.

\begin{table}[t]\footnotesize
    \centering
    \caption{\textbf{The description of real-world workloads.}}
    \vspace{-8pt}
    \begin{tabular}{m{1.4cm} m{6.5cm}}
        \hline
        \multicolumn{1}{c}{\textbf{Workload}} &
          \multicolumn{1}{c}{\textbf{Description}} \\ \hline
        RAG &
          An on-device smart assistant that personalizes its model with user data. It combines vector and keyword searches, using a reranker to select the optimal final result. \\ \hline
        \begin{tabular}[c]{@{}l@{}}Agent\\ Memory~\cite{zhang2025mobiagent}\end{tabular} &
          An on-device agent leverages a reranker in its agent memory to cache actions, reducing costly model generations. \\ \hline
        \begin{tabular}[c]{@{}l@{}}LLM Long\\ Context\\Selection~\cite{jiang2024longllmlingua}\end{tabular} &
          For on-device deployment of LLM handling extended contexts, a top-$K$ selection mechanism is employed to identify the most related contextual segments, conforming to the model's finite context window limitations. \\ \hline
    \end{tabular}
    \vspace{-10pt}
    \label{tab:realworld-workload-desc}
\end{table}

\paragraph{Metrics.}
Our evaluation focuses on the following metrics:
\begin{itemize}[leftmargin=*]
    \item \textbf{Latency.} We measure the inference latency of the reranking models.
    \item \textbf{Precision.} We employ Precision@$K$ to evaluate the model precision. Precision@$K$ measures the ratio between the number of relevant items contained in the top-$K$ results and $K$. When the ground truth is less than $K$, we take the ratio between the number of relevant items contained in the top-$K$ and the number of ground truth.
    \item \textbf{Memory footprint.} We focus on both the mean and the peak memory footprint of the model in inference.
\end{itemize}

\subsection{Microbenchmarks}
In microbenchmarks, we extensively evaluate the latency, precision, and memory footprint of compared systems.

\newcommand{\numpm}[2]{#1\,/\,{-#2}}
\newcommand{\fn}[1]{{\footnotesize #1}}
\begin{table*}[t]
    \centering
    \small
    \setlength{\tabcolsep}{4pt}
    \renewcommand{\arraystretch}{1.2}
    \begin{tabular}{@{}cl c cc cc cc@{}}
    \toprule
    \multirow{2}{*}{\textbf{Model}} & 
    \multirow{2}{*}{\textbf{System}} & 
    \multirow{2}{*}{\textbf{Baseline}} &
    \multicolumn{2}{c}{\textbf{Precision@1}} &
    \multicolumn{2}{c}{\textbf{Precision@5}} &
    \multicolumn{2}{c}{\textbf{Precision@10}} \\
    \cmidrule(lr){4-5} \cmidrule(lr){6-7} \cmidrule(lr){8-9}
    & & &
    \makecell{Lat. Reduction\\[-2pt]\scriptsize Range (Mean)} &
    \makecell{Prec. Loss\\[-2pt]\scriptsize Mean\,/\,Max} &
    \makecell{Lat. Reduction\\[-2pt]\scriptsize Range (Mean)} &
    \makecell{Prec. Loss\\[-2pt]\scriptsize Mean\,/\,Max} &
    \makecell{Lat. Reduction\\[-2pt]\scriptsize Range (Mean)} &
    \makecell{Prec. Loss\\[-2pt]\scriptsize Mean\,/\,Max} \\
    \midrule
    \multirow{3}{*}{\makecell{Qwen3\\[-1pt]0.6B}}
     & \multirow{2}{*}{\sys} & HF &
    \fn{10.5--53.9\% (33.1\%)} & \multirow{2}{*}{\fn{\numpm{0.003}{0.003}}} &
    \fn{9.0--51.7\% (32.9\%)} & \multirow{2}{*}{\fn{\numpm{0.002}{0.003}}} &
    \fn{8.5--52.3\% (32.2\%)} & \multirow{2}{*}{\fn{\numpm{0.000}{0.003}}} \\
     &  & HF Offload &
    \fn{32.2--67.4\% (47.6\%)} & &
    \fn{32.5--66.2\% (47.6\%)} & &
    \fn{32.0--65.8\% (47.1\%)} & \\
    \cmidrule(l){2-9}
     & {\sys} Quant & HF Quant &
    \fn{5.6--50.1\% (20.3\%)} & \fn{\numpm{0.001}{0.001}} &
    \fn{5.7--48.0\% (18.3\%)} & \fn{\numpm{0.002}{0.004}} &
    \fn{5.3--48.6\% (19.6\%)} & \fn{\numpm{0.000}{0.004}} \\
    \midrule
    \multirow{3}{*}{\makecell{Qwen3\\[-1pt]4B}}
     & \multirow{2}{*}{\sys} & HF &
    \multicolumn{1}{c}{\fn{\textsf{OOM}}} & \multirow{2}{*}{\fn{\numpm{-0.001}{0.003}}} &
    \multicolumn{1}{c}{\fn{\textsf{OOM}}} & \multirow{2}{*}{\fn{\numpm{-0.001}{0.005}}} &
    \multicolumn{1}{c}{\fn{\textsf{OOM}}} & \multirow{2}{*}{\fn{\numpm{0.000}{0.004}}} \\
     &  & HF Offload &
    \fn{12.8--65.7\% (38.2\%)} & &
    \fn{10.7--65.4\% (37.7\%)} & &
    \fn{6.2--64.1\% (35.3\%)} & \\
    \cmidrule(l){2-9}
     & {\sys} Quant & HF Quant &
    \fn{5.1--43.5\% (18.7\%)} & \fn{\numpm{0.000}{0.001}} &
    \fn{5.2--41.8\% (17.9\%)} & \fn{\numpm{0.000}{0.003}} &
    \fn{5.7--40.7\% (17.1\%)} & \fn{\numpm{0.000}{0.004}} \\
    \midrule
    \multirow{3}{*}{\makecell{Qwen3\\[-1pt]8B}}
     & \multirow{2}{*}{\sys} & HF &
    \multicolumn{1}{c}{\fn{\textsf{OOM}}} & \multirow{2}{*}{\fn{\numpm{0.039}{0.001}}} &
    \multicolumn{1}{c}{\fn{\textsf{OOM}}} & \multirow{2}{*}{\fn{\numpm{0.048}{0.004}}} &
    \multicolumn{1}{c}{\fn{\textsf{OOM}}} & \multirow{2}{*}{\fn{\numpm{0.025}{0.006}}} \\
     &  & HF Offload &
    \fn{23.2--80.7\% (54.6\%)} & &
    \fn{23.4--80.5\% (53.6\%)} & &
    \fn{23.5--80.1\% (53.3\%)} & \\
    \cmidrule(l){2-9}
     & {\sys} Quant & HF Quant &
    \fn{6.5--64.3\% (36.8\%)} & \fn{\numpm{0.040}{0.006}} &
    \fn{7.1--62.6\% (36.1\%)} & \fn{\numpm{0.056}{0.001}} &
    \fn{9.6--62.2\% (36.0\%)} & \fn{\numpm{0.031}{0.008}} \\
    \midrule
    \multirow{3}{*}{\makecell{Bge\\[-1pt]M3}}
     & \multirow{2}{*}{\sys} & HF &
    \fn{8.1--41.6\% (24.6\%)} & \multirow{2}{*}{\fn{\numpm{0.000}{0.005}}} &
    \fn{6.2--41.7\% (22.4\%)} & \multirow{2}{*}{\fn{\numpm{0.000}{0.004}}} &
    \fn{4.5--43.2\% (20.5\%)} & \multirow{2}{*}{\fn{\numpm{0.002}{0.007}}} \\
     &  & HF Offload &
    \fn{52.4--80.0\% (70.9\%)} & &
    \fn{50.9--79.4\% (70.1\%)} & &
    \fn{49.9--79.7\% (69.3\%)} & \\
    \cmidrule(l){2-9}
     & {\sys} Quant & HF Quant &
    \fn{4.3--44.9\% (25.6\%)} & \fn{\numpm{0.002}{0.006}} &
    \fn{1.3--41.6\% (22.5\%)} & \fn{\numpm{0.004}{0.006}} &
    \fn{0.5--41.8\% (21.1\%)} & \fn{\numpm{0.003}{0.006}} \\
    \midrule
    \multirow{3}{*}{\makecell{Bge\\[-1pt]MiniCPM}}
     & \multirow{2}{*}{\sys} & HF &
    \fn{9.2--72.8\% (44.3\%)} & \multirow{2}{*}{\fn{\numpm{0.000}{0.003}}} &
    \fn{6.1--66.9\% (35.5\%)} & \multirow{2}{*}{\fn{\numpm{-0.003}{0.006}}} &
    \fn{5.6--62.4\% (29.6\%)} & \multirow{2}{*}{\fn{\numpm{-0.002}{0.005}}} \\
     &  & HF Offload &
    \fn{13.7--89.2\% (56.7\%)} & &
    \fn{14.2--87.0\% (50.7\%)} & &
    \fn{15.4--84.3\% (47.6\%)} & \\
    \cmidrule(l){2-9}
     & {\sys} Quant & HF Quant &
    \fn{6.1--72.2\% (36.7\%)} & \fn{\numpm{0.000}{0.003}} &
    \fn{3.2--66.8\% (28.2\%)} & \fn{\numpm{-0.002}{0.006}} &
    \fn{2.3--59.1\% (23.7\%)} & \fn{\numpm{-0.002}{0.006}} \\
    \bottomrule
    \end{tabular}
    \vspace{2pt}
    \caption{\textbf{Summary of latency and precision evaluation on 5 models, 2 platforms, and 18 datasets.}
    For each precision@$K$, we report our system's mean latency reduction range (with mean) and mean/max precision loss compared to baselines across 18 datasets $\times$ 2 platforms.}
    \vspace{-15pt}
    \label{tab:microbenchmark-latency-summary}
\end{table*}

\begin{figure*}[hbt]
    \centering
    \includegraphics[width=\linewidth]{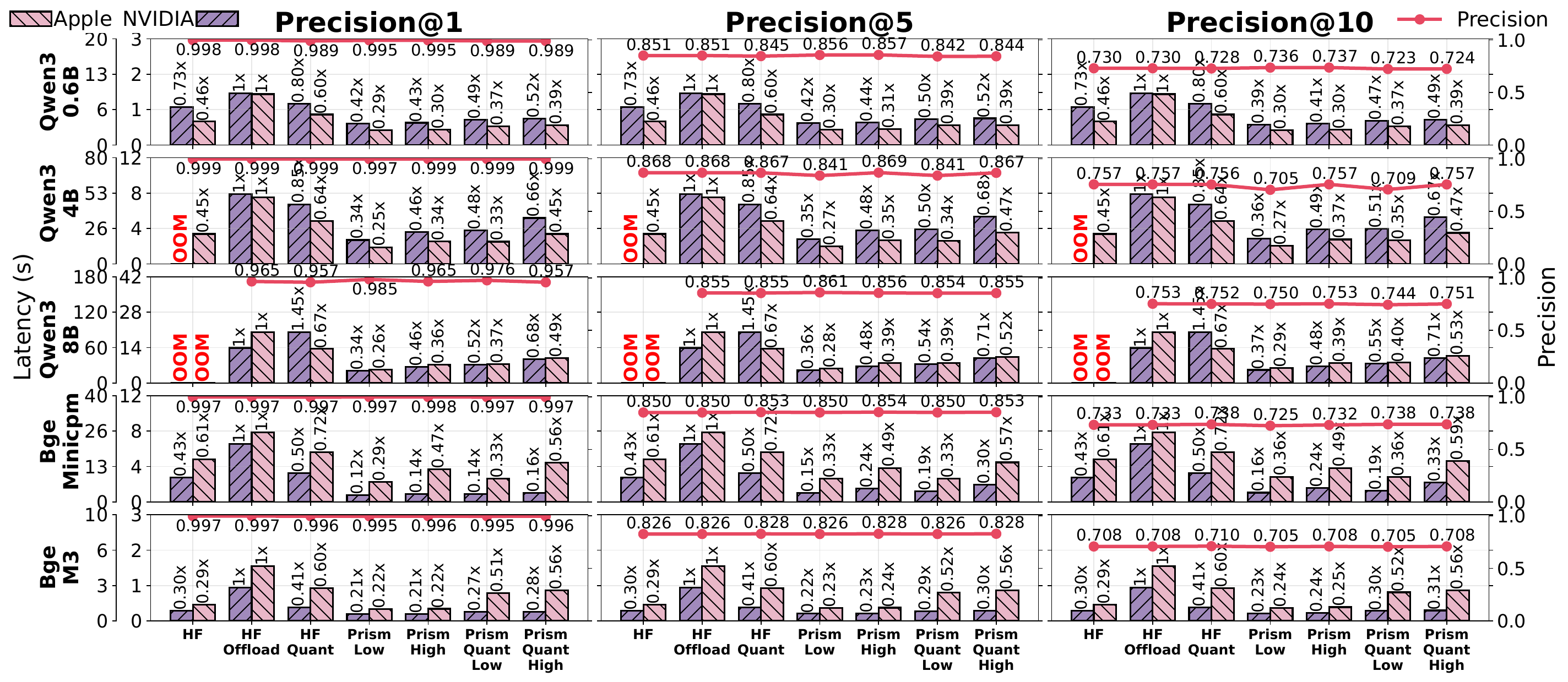}
    \vspace{-24pt}
    \caption{\textbf{The detailed latency and precision evaluation on the Wikipedia dataset~\cite{ellamind2023wiki}.}}
    \vspace{-10pt}
    \label{fig:microbenchmark-latency-wikipedia}
\end{figure*}

\begin{figure*}[bt]
    \centering
    \includegraphics[width=0.95\linewidth]{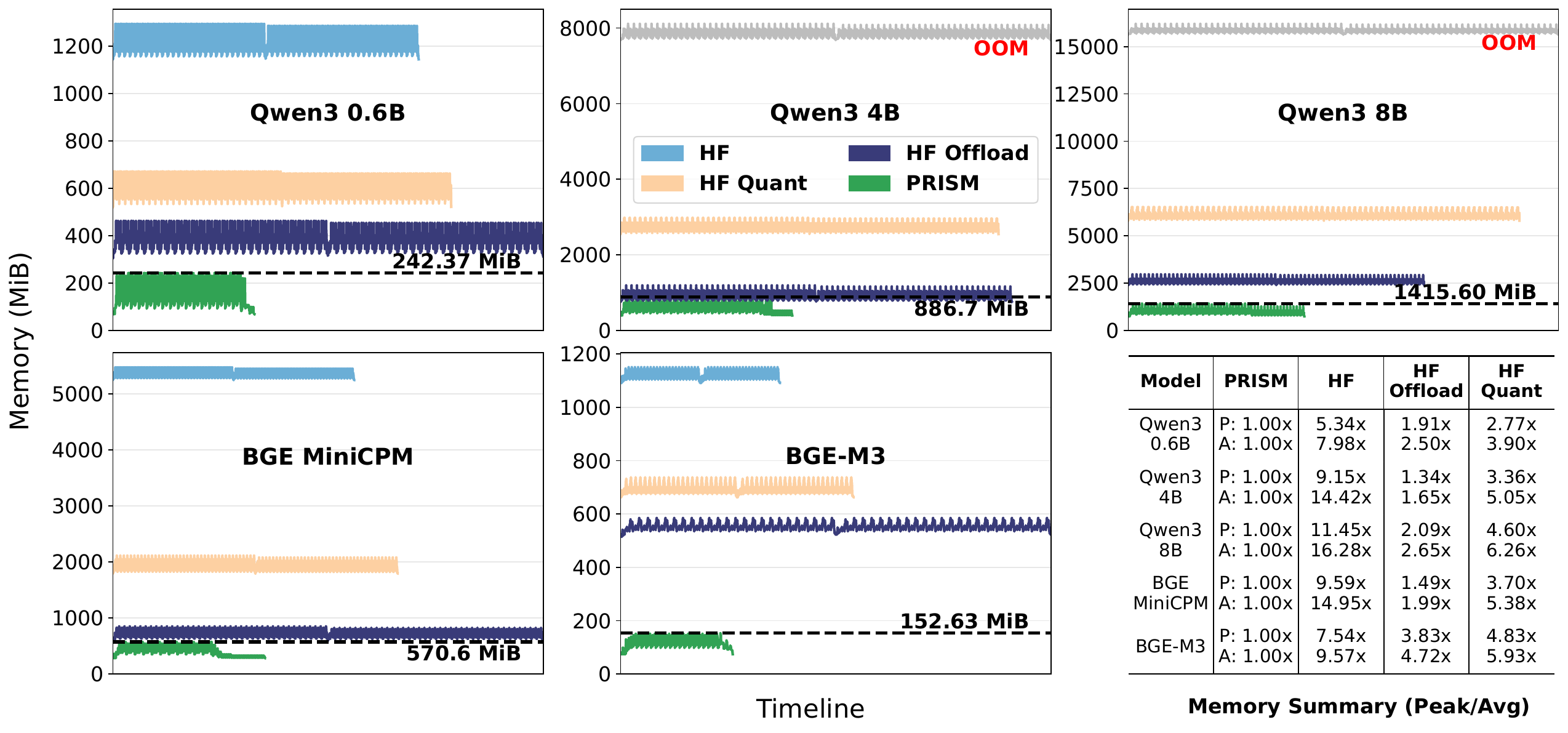}
    \vspace{-12pt}
    \caption{\textbf{The memory footprint in microbenchmarks.}}
    \vspace{-10pt}
    \label{fig:microbenchmark-memory}
\end{figure*}

\begin{figure*}[hbt]
    \centering
    \includegraphics[width=0.9\linewidth]{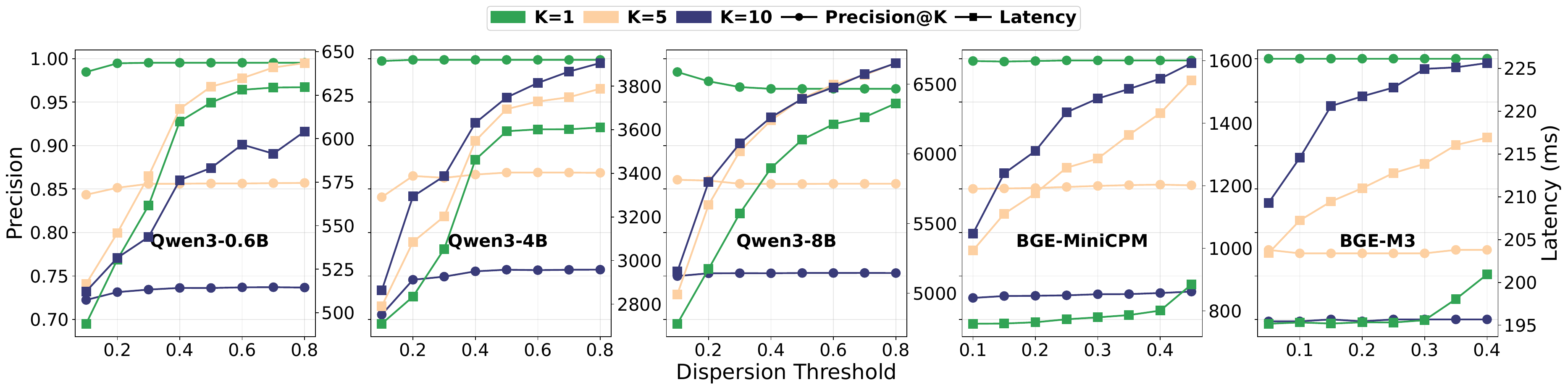}
    \vspace{-12pt}
    \caption{\textbf{Tuning the latency and precision trade-off.}}
    \vspace{-10pt}
    \label{fig:ablation-latency-precision-tradeoff}
\end{figure*}

\paragraph{Latency and precision.}
\autoref{tab:microbenchmark-latency-summary} summarizes our latency and precision evaluation on five models across 18 datasets and two platforms, reporting the mean latency reduction and the maximum precision loss on Precision@$K \in \{1, 5, 10\}$ out of 20 candidates compared to the baselines.
\autoref{fig:microbenchmark-latency-wikipedia} further zooms in the latency and precision results on the Wikipedia dataset~\cite{ellamind2023wiki} across NVIDIA and Apple platforms.
Each subplot corresponds to a specific model and top-$K$ configuration, showing latency (left y-axis) and precision (right y-axis).
Bars represent latency (purple: NVIDIA; pink: Apple), and the number atop each bar indicates the speedup relative to the HF Offload baseline.
The line plot reports Precision@$K$.
As precision is platform-independent, both platforms share a single point.
In \autoref{fig:microbenchmark-latency-wikipedia}, we additionally plot {\sys}/{\sys} Quant under both low and high dispersion thresholds (see~\autoref{sec:progress-cluster-pruning}) to show the threshold's configurability.

Overall, \autoref{tab:microbenchmark-latency-summary} shows that our systems deliver substantial mean latency reductions while preserving precision, strongly validating the effectiveness of our design.
Quantitatively, our system's benefits are significant.
{\sys} reduces latency up to 89.2\% over HF Offload at Precision@1 on \texttt{Bge-MiniCPM}, while the maximum precision loss stays within -0.003.
{\sys} Quant also provides meaningful speedups over HF Quant (e.g., up to 72.2\% at Precision@1), with the maximum precision loss bounded by -0.003.
For larger models such as \texttt{Qwen3-4B/8B}, the HF baseline fails to run on our hardware platforms due to its large memory footprint (OOM).
In contrast, our system enables low-latency inference for these powerful models, further underscoring its practical utility and effectiveness.

Zooming in on the Wikipedia dataset, \autoref{fig:microbenchmark-latency-wikipedia} shows the detailed latency and precision on the compared systems and two platforms.
{\sys} consistently achieves the lowest latency with high precision, followed closely by {\sys} Quant, which provides the next-best performance.
Furthermore, the results highlight a clear trade-off configurable via the dispersion threshold: increasing the threshold from Low to High improves precision at the cost of a smaller latency reduction.
Quantitatively, {\sys} reduces latency up to 72\% compared to HF and 88\% compared to HF Offload, with no loss in precision.
For the \texttt{Qwen3-8B} model, the {\sys} with Low threshold setting abnormally increases precision significantly compared to the HF baselines.
We attribute this to the overfitting of the \texttt{Qwen3-8B} model~\cite{zhang2025qwen3embedding}; our low-threshold {\sys} provides a regularizing effect by bypassing the later layers, thereby enhancing its generalization.

In summary, extensive latency microbenchmarks demonstrate our system achieves substantial latency reductions (up to 89.2\%) while preserving precision.
This effectiveness holds across a range of models, datasets, and hardware platforms.
Crucially, our system enables low-latency inference on large models that are otherwise infeasible to run.

\paragraph{Memory Footprint.}
\autoref{fig:microbenchmark-memory} illustrates the inference memory footprint overtime of the compared systems across five different models.
The benchmark was conducted on the NVIDIA platform with ranking top-10 out of 20 input candidates with an average sequence length of 500.
The results on the Apple platform are similar to those on the NVIDIA platform, we do not elaborate further.

In each subfigure, the x-axis represents the timeline and the y-axis shows memory usage.
Each line terminates upon inference completion, thus its length indicates the inference latency.
We annotate the peak memory of {\sys} in each subfigure and present the peak and average memory statistics in the table at the bottom right.
To demonstrate the memory footprint of HF on \texttt{Qwen3-4B} and \texttt{Qwen3-8B} that cannot run in the NVIDIA platform due to the OOM error, we measure them on an NVIDIA A800 GPU.
Therefore, the lengths of their corresponding lines do not represent valid latencies.
Besides, we omit the curve for {\sys} Quant because it nearly overlaps with that of {\sys}, enhancing visual clarity.

Overall, our system achieves the lowest memory footprint among all baselines while simultaneously delivering the lowest latency. 
This result demonstrates that our technique enables models to run faster with substantially less memory, a dual benefit not offered by competing approaches. 
In contrast, other memory-saving baselines like HF Offload and HF Quant trade latency for lower memory.
Quantitatively, our system reduces peak memory by $5.34\times$ -- $11.45\times$ compared to HF, $1.34\times$ -- $3.83\times$ compared to HF Offload, and $2.77\times$ -- $4.83\times$ compared to Quant.
This substantial memory saving is particularly critical for resource-constrained edge devices, as it alleviates memory pressure and enables robust co-location of multiple applications.

In summary, our system substantially reduces both memory footprint and inference latency.
This unique combination of memory efficiency and high performance is crucial for resource-constrained edge devices.

\paragraph{Tuning the latency-precision trade-off.}
~\autoref{fig:ablation-latency-precision-tradeoff} demonstrates our system's ability to navigate the latency-precision spectrum by tuning the dispersion threshold. 
We evaluate this capability on five models under Precision@1/5/10.
As a general trend, increasing the threshold improves precision at the cost of higher latency.
This result validates that our system can be configured to operate at different points on the latency-precision spectrum to meet diverse application requirements.
Notably, \texttt{Qwen3-8B} exhibits an inverse trend, achieving peak precision at the lowest threshold.
We attribute this to the overfitting of this model, which is also shown in the official benchmark~\cite{qwen2025benchmark}; the pruning mechanism acts as a form of regularization, mitigating this effect.

\begin{figure*}[hbt]
    \centering
    \begin{subfigure}{0.33\linewidth}
        \includegraphics[width=\linewidth]{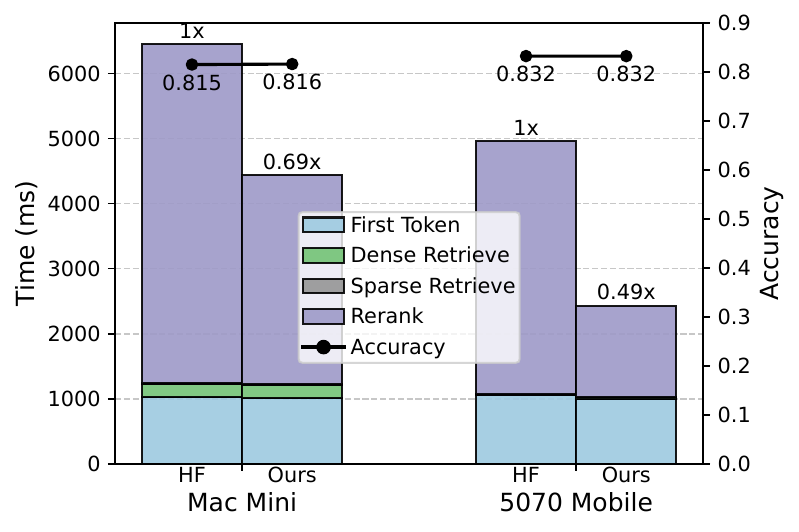}
        \vspace{-20pt}
        \caption{\textbf{Latency \& precision of RAG.}}
        \label{fig:macro_rag_latency}
    \end{subfigure}
    \begin{subfigure}{0.33\linewidth}
        \centering
        \includegraphics[width=\linewidth]{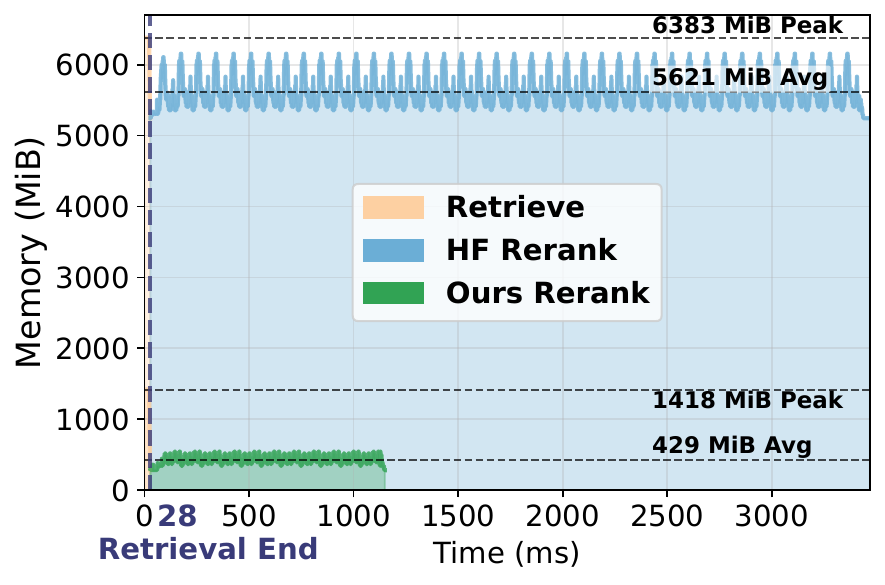}
        \vspace{-20pt}
        \caption{\textbf{Memory footprint on NVIDIA.}}
        \label{fig:macro_rag_memory_5070}
    \end{subfigure}
    \begin{subfigure}{0.33\linewidth}
        \centering
        \includegraphics[width=\linewidth]{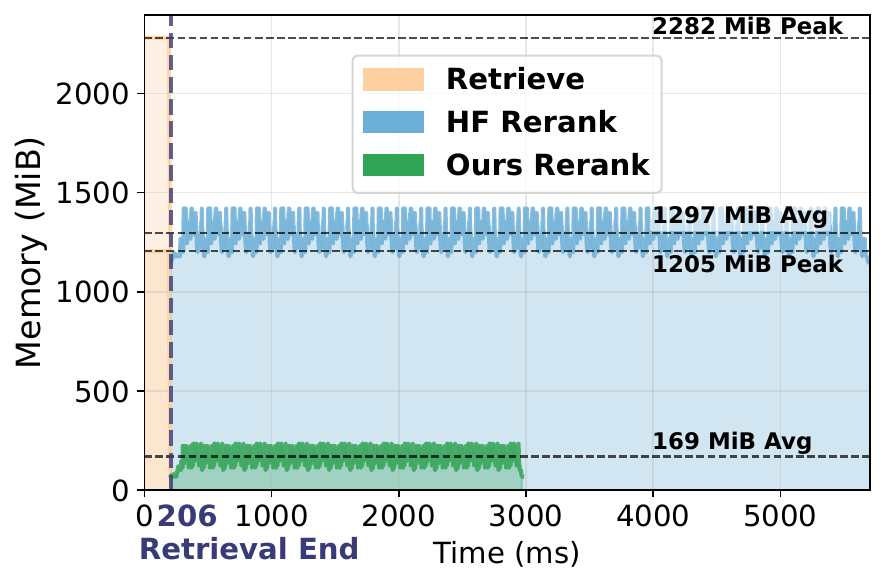}
        \vspace{-20pt}
        \caption{\textbf{Memory footprint on Apple.}}
        \label{fig:macro_rag_memory_macmini}
    \end{subfigure}
    \vspace{-20pt}
    \caption{\textbf{The latency, precision, and memory footprint of RAG.}}
    \label{fig:macro_rag_memory}
\end{figure*}

\begin{figure*}[hbt]
    \centering
    \begin{minipage}[t]{0.33\linewidth}
        \vspace*{0pt}
        \centering
        \includegraphics[width=\linewidth]{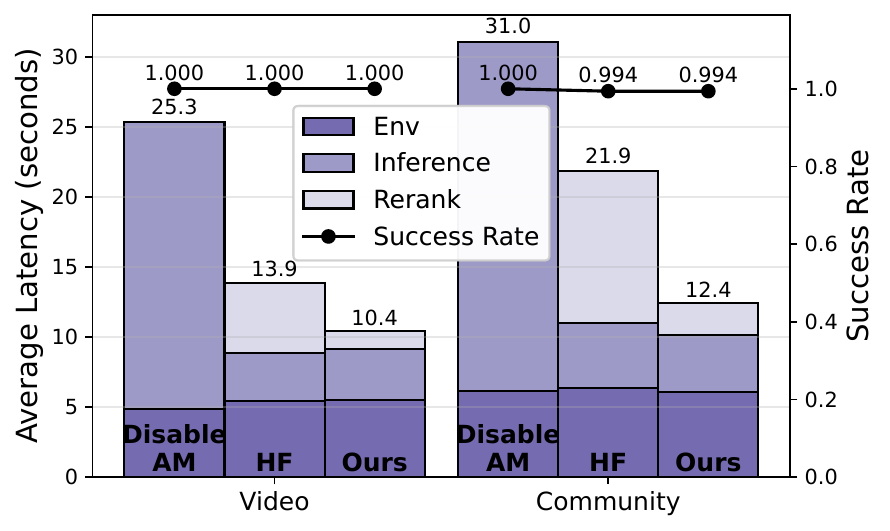}
        \vspace{-20pt}
        \caption{\textbf{Latency \& precision of AM.}}
        \label{fig:macro_agent_latency}
    \end{minipage}%
    \begin{minipage}[t]{0.33\linewidth}
        \vspace*{0pt}
        \centering
        \includegraphics[width=0.96\linewidth]{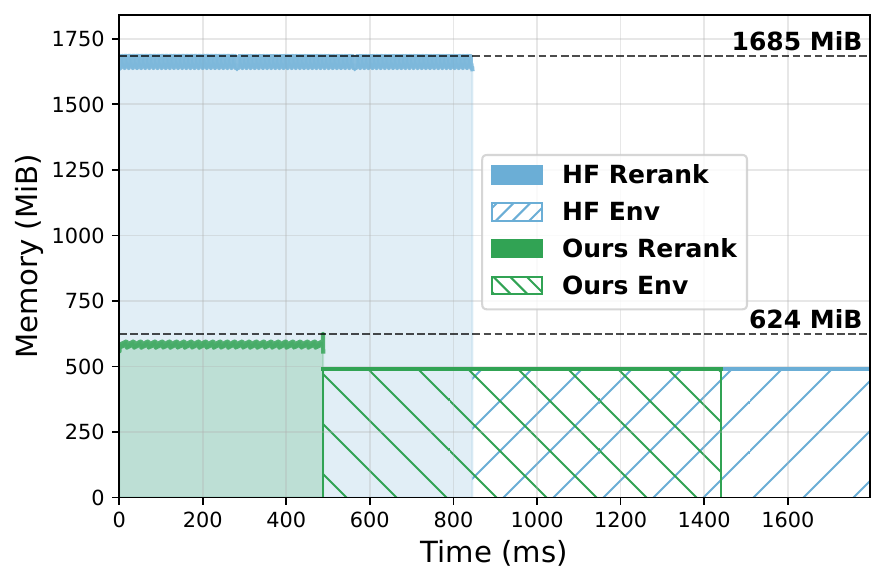}
        \vspace{-14pt}
        \caption{\textbf{Memory footprint of AM.}}
        \label{fig:macro_agent_memory}
    \end{minipage}%
    \begin{minipage}[t]{0.33\linewidth}
        \vspace*{0pt}
        \centering
        \includegraphics[width=\linewidth]{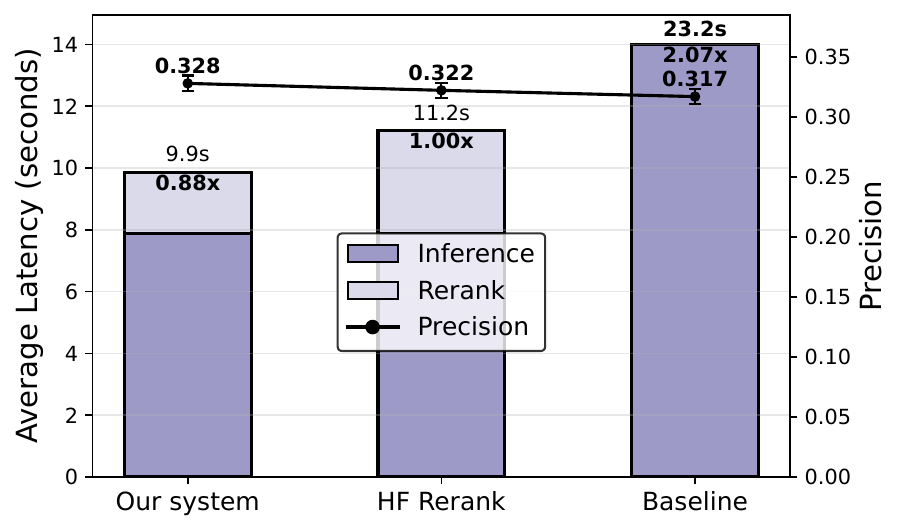}
        \vspace{-19pt}
        \caption{\textbf{Latency \& precision of LCS.}}
        \label{fig:macro_long_context_selection_latency}
    \end{minipage}
    \vspace{-10pt}
\end{figure*}

\subsection{Real-world Evaluations}
\paragraph{Retrieval-Augmented Generation (RAG).}
We evaluate our system on the RAG-based personal assistant scenario.
During an offline indexing phase, user's personal data is converted into vector embeddings by the embedding model and stored in the vector database.
When a user query arrives, we perform a hybrid search using both dense retrieval (i.e., vector search) and sparse retrieval (i.e., keyword search) to find the top-10 relevant documents, respectively.
Then, a reranking model consolidates the results and selects the top-10 documents, which are then sent to an LLM for generation.

We employ the DiskANN-based Milvus~\cite{Milvus, Wang2021milvus} as our vector database, the \texttt{Qwen3-Embedding-0.6B} for embedding.
For reranking, we employ \texttt{Qwen3-Reranker-0.6B} on the Apple platform and \texttt{Bge-Minicpm} on the NVIDIA platform.
For generation, we deploy a \texttt{Qwen3-32B} model on a server with two NVIDIA A800 GPU.

\autoref{fig:macro_rag_latency} compares the latency and precision of HF and {\sys}.
Our system achieves significant performance gains, reducing latency by 51\% on NVIDIA platform and 31\% on Apple platform, respectively.
Crucially, these improvements come with almost no loss in model precision.
\autoref{fig:macro_rag_memory_5070} and~\autoref{fig:macro_rag_memory_macmini} shows the memory footprint of HF and {\sys}.
Our system also substantially lowers peak memory by up to 77.8\% and average memory by 92.3\%. 
This large reduction in average memory stems from the aggressive memory optimization of the reranking phase, which dominates the overall execution time.

\paragraph{Agent Memory (AM).}
We evaluate our system in an agent memory application~\cite{zhang2025mobiagent}.
This application optimizes GUI-based agent by caching past successful action trajectories to bypass expensive and redundant Vision-Language Model (VLM)~\cite{mobiagent2025vlm} inference.
The core of the agent memory lies in selecting the most semantic relevant trajectories~\cite{mobiagent2025trace}, which is performed by a reranker.
We employ \texttt{Qwen3-Reranker-0.6B} reranker and evaluate our system in the NVIDIA platform, the VLM is serving on two NVIDIA A800 server.
The test results on the Apple platform are similar to those on the NVIDIA platform, so we will not elaborate further.

As shown in~\autoref{fig:macro_agent_latency}, we evaluate the task completion latency and the task success rate in two different workloads.
Our system significantly reduces the latency by 25.2\% in the video workload and 43.4\% in the community scenario, respectively.
Crucially, these improvements come with no loss in the task success rate.
\autoref{fig:macro_agent_memory} shows the memory footprint in the period of a single click action performing by the agent.
Compared to HF, {\sys} reduces the peak memory usage by 63.0\%.
Such substantial savings are particularly valuable on resource-constrained edge devices, where lower memory usage directly increases the keep-alive rate of applications and improves the users' experience.

\begin{figure}[t]
    \centering
    \includegraphics[width=0.75\linewidth]{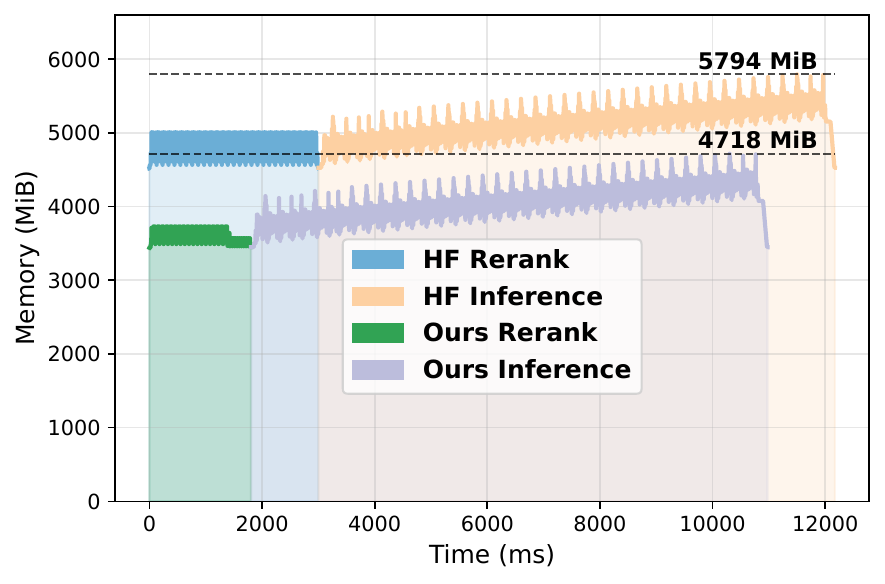}
    \vspace{-10pt}
    \caption{\textbf{Memory footprint of LCS.}}
    \vspace{-20pt}
    \label{fig:macro_long_context_selection_memory}
\end{figure}

\paragraph{LLM Long Context Selection (LCS).}
LLM long context selection aims to select the most relevant information in an ultra long context, thereby accelerating inference.
In this real-world scenario, we employ a \texttt{Qwen3-Reranker-0.6B} reranker to select the most relevant information and then feed them to a quantized \texttt{Qwen3-4B-Instruct} for generation.
The evaluations are conducted on the NVIDIA platform with the LongBench2~\cite{bai2024longbench2} benchmark.
We compare the latency and precision of three systems: {\sys}, HF Reranker, and No Reranker.

\autoref{fig:macro_long_context_selection_latency} reports the end-to-end latency and precision.
Generally, {\sys} achieves a latency reduction of 11.6\% compared to HF Reranker and 57.3\% compared to No Reranker, with even marginal precision increasing.
For precision, the two systems with reranker surpass the No Reranker, which is distracted by the irrelevant information.
Besides, the slight precision gain observed in {\sys} over HF Reranker fall within normal variation.
\autoref{fig:macro_long_context_selection_memory} compares the memory footprint of HF Reranker and {\sys} in one generation.
{\sys} reduces peak memory by about 1\,GiB compared to the HF Reranker.

In summary, {\sys} consistently outperforms the baselines in latency and memory footprint.

\begin{figure}[hbt]
    \centering
    \vspace{-10pt}
    \includegraphics[width=\linewidth]{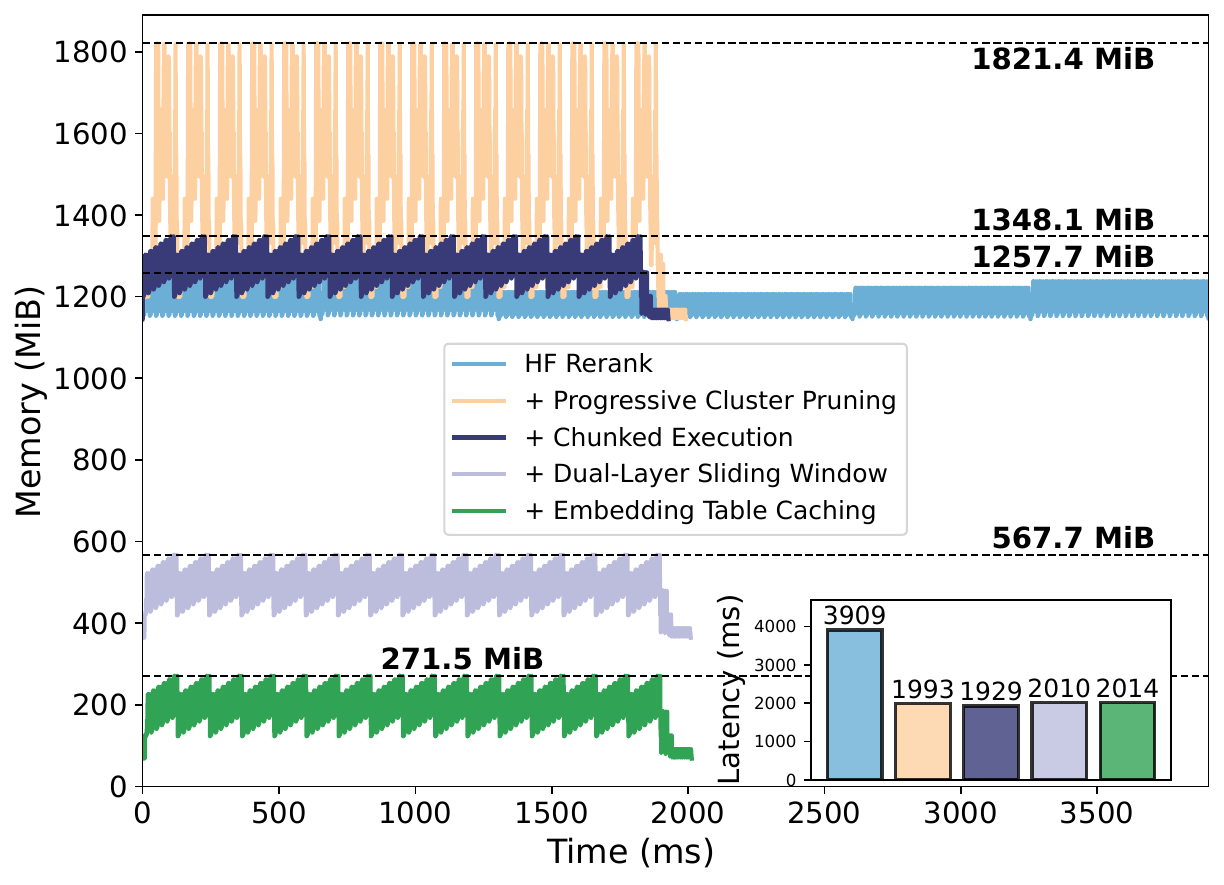}
    \vspace{-25pt}
    \caption{\textbf{Memory \& latency ablation of four techniques.}}
    \vspace{-10pt}
    \label{fig:ablation_memory_latency}
\end{figure}
\subsection{Ablation Study}
\label{sec:abalation}
We conduct an ablation study to show the contributions of our four proposed techniques by applying them incrementally.
We measure the latency and memory footprint running the \texttt{Qwen3-Reranker-0.6B} on the NVIDIA platform to rank 60 candidates with an average length of 500. 

\autoref{fig:ablation_memory_latency} illustrates how memory footprint and latency change as each technique is applied incrementally. 
Starting from the baseline, we first apply progressive cluster pruning.
It reduces the latency by 49.0\%, but increases the peak memory by 44.8\% due to the monolithic batch.
Then, we apply the chunked execution to reduce this memory overhead to 7.2\%.
The remaining overhead stems from storing hidden states for all chunks, a requirement for the monolithic forwarding scheme.
Next, we apply the overlapped layer streaming, significantly reduces memory usage by 57.8\%. 
This optimization incurs a modest 81\,ms latency overhead because the reduced computation time from pruning no longer fully hides the I/O latency.
Finally, embedding table caching eliminates the last dominant memory bottleneck, reducing the peak memory usage to 271\,MiB with a negligible 4\,ms latency overhead.
When combined, {\sys} achieves a 78.4\% reduction in peak memory and a 48.5\% reduction in latency compared to the baseline, demonstrating the effectiveness of our techniques when working in concert. 

These results validate that each technique successfully achieves its intended optimization goal, and their integration yields a system optimized for both memory and latency.


\section{Discussion}

\paragraph{Flexibility for diverse application needs.}
{\sys}'s design provides the flexibility to cater to diverse application requirements.
In most scenarios, such as selecting documents for a RAG pipeline, the primary goal is to identify the top-$K$ candidates regardless of their precise internal ranking.
For such cases, the pruning both winners and hopeless candidates can maximize latency reduction as demonstrated in our evaluation.
Furthermore, {\sys} is equally capable of handling applications where the exact rank order or the final scores are critical.
{\sys} supports this by only pruning the hopeless candidates while allowing the top contenders to undergo full inference.
This adaptability allows developers to tune the system for their specific latency budget and application-level quality requirements.

\paragraph{Generality beyond evaluated models.}
While our evaluation focused on a representative set of state-of-the-art rerankers, we have also observed the core insight of sequence-level sparsity is a general characteristic of cross-encoder architectures.
The hierarchical nature of transformers, where earlier layers capture broader contextual features and later layers refine nuanced semantic relationships, naturally leads to the progressive emergence of stable relative rankings.
Our preliminary experiments with other cross-encoder models, including LLM as rerankers (e.g., \texttt{Qwen3-4B-Instruct}), confirm this pattern.
This suggests that the principles behind {\sys} are not limited to specialized reranker models but can likely be extended to a broader class of transformer-based models performing semantic selection tasks.

\paragraph{Discussion on related work.}
For related works on model compression, {\sys} is a training-free system and is orthogonal to a wide array of model compression techniques.
As shown by our ``{\sys} Quant'' evaluation, its benefits seamlessly compound with post-training quantization methods~\cite{frantar2023gptq,lin2024awq}.
Moreover, our approach can be readily applied to models that have already undergone training-based compression~\cite{ma2023llmpruner,sanh2020distilbert,chen2024efficientqat}.
This orthogonality allows {\sys} to be integrated with existing and future model compression advancements to further push the efficiency frontier of on-device AI.
For related works on memory optimization, PrefillOnly~\cite{du2025prefillonly} proposes chunking batches to reduce the memory overhead of intermediate tensors.
{\sys} additionally proposes dynamic offloading of hidden states to further reduce peak memory. 
Furthermore, tailored for edge scenarios, {\sys} introduces overlapped layer streaming and embedding table cache, significantly reducing the memory footprint of model weights.


\section{Conclusion}

We introduce {\sys}, a training-free inference system that re-frames reranking to focus on relative rankings, enabling a highly efficient monolithic forwarding architecture.
Our system uses progressive cluster pruning and a series of memory optimizations to significantly reduce latency by up to 89.2\% and peak memory by up to 91.3\%, substantially advancing on-device semantic selection.


\section*{Acknowledgments}
We sincerely thank our shepherd, Supawit Chockchowwat, and the anonymous reviewers for their insightful comments.
This work is supported in part by the National Natural Science Foundation of China (No. 62132014), the Fundamental Research Funds for the Central Universities, Fundamental and Interdisciplinary Disciplines Breakthrough Plan of the Ministry of Education of China (JYB2025XDXM113), and Huawei Technologies.
The corresponding authors
are Mingkai Dong (mingkaidong@sjtu.edu.cn) and Dingji Li\\ (lidingji1997@hotmail.com).

\bibliographystyle{ACM-Reference-Format} 
\bibliography{refs}

\clearpage
%

\appendix
\section{Artifact Appendix}

\subsection{Abstract}
The artifact includes the source code of {\sys} and the baselines, and the scripts to reproduce the results in the paper.

\subsection{Description \& Requirements}

\subsubsection{How to access}
The artifact is available at: \url{https://ipads.se.sjtu.edu.cn:1312/opensource/monolithic_forwarding} and archived at \url{https://doi.org/10.5281/zenodo.18809731}.

\subsubsection{Hardware dependencies}
\begin{itemize}[noitemsep]
    \item \textbf{GPU}: NVIDIA GPU with at least 8\,GB VRAM or M-series Apple Silicon.
    \item \textbf{RAM}: 16\,GB minimum.
    \item \textbf{Disk}: $\sim$50\,GB for datasets and model checkpoints.
\end{itemize}
Experiments in the paper use an NVIDIA RTX 5070 Laptop (8\,GiB) and an Apple Mac Mini M2 (16\,GiB unified memory). Reproduction on equivalent or better NVIDIA hardware is supported; Apple platform results can be reproduced on M-series Macs with sufficient memory.

\subsubsection{Software dependencies}
\begin{itemize}[noitemsep]
    \item \textbf{OS}: Linux (tested on Ubuntu 22.04) and macOS (tested on Sequoia 15.1).
    \item \textbf{Environment}: Conda (Miniconda or Anaconda).
    \item \textbf{CUDA}: 12.1 or newer (for NVIDIA).
    \item \textbf{Docker}: Required for Milvus-based RAG experiments (Section~6.3).
\end{itemize}

\subsubsection{Benchmarks}
\begin{itemize}[noitemsep]
    \item \textbf{Microbenchmarks (Section~6.2)}: 18 datasets and reranker models (Qwen3-Reranker-0.6B/4B/8B, Bge-Reranker-v2-MiniCPM, Bge-Reranker-v2-M3) downloaded via provided \texttt{download\_models.sh} from HuggingFace.
    \item \textbf{Real-world (Section~6.3)}: RAG pipeline (with Milvus); Agent Memory (video and community scenarios); Long Context Selection (LongBench-style workloads). Data and workloads are included or fetched by the artifact scripts.
\end{itemize}

\subsection{Set-up}

\begin{enumerate}[noitemsep]
    \item Clone the artifact repository and enter its root directory.
    \item Run \texttt{bash install\_dependencies.sh} to create the Conda environment, install Python dependencies, and build C extensions.
    \item Run \texttt{bash download\_models.sh} to fetch model checkpoints from HuggingFace.
    \item Run \texttt{bash run\_demo.sh} to perform kick-the-tires verification with a small example. Successful completion indicates the environment is ready.
\end{enumerate}
For detailed steps, see \texttt{quickstart.md} in the repository.

\subsection{Evaluation workflow}
\subsubsection{Major Claims}

\begin{itemize}[leftmargin=*]
    \item \textbf{(C1) Microbenchmark latency:} {\sys} reduces latency (e.g., up to 89.2\%) versus HF vanilla/offload/quant without compromising precision. Supported by experiments (E1.1); results in Section~6.2 and \textbf{Figure~8}.
    \item \textbf{(C2) Microbenchmark memory:} {\sys} reduces peak GPU memory by 5.34$\times$--11.45$\times$ vs.\ HF, 1.34$\times$--3.83$\times$ vs.\ HF Offload, and 2.77$\times$--4.83$\times$ vs.\ Quant. Supported by (E1.2); results in Section~6.2 and \textbf{Figure~9}.
    \item \textbf{(C3) Latency--precision trade-off:} Increasing the threshold improves precision at the cost of higher latency. Supported by (E1.3); results in Section~6.2 and \textbf{Figure~10}.
    \item \textbf{(C4) RAG pipeline:} {\sys} reduces latency by 51\% on NVIDIA and 31\% on Apple, and reduces peak memory by 77.8\% and average memory by 92.3\%, with no precision loss. Supported by (E2.1); results in Section~6.3 and \textbf{Figure~11}.
    \item \textbf{(C5) Agent Memory:} {\sys} reduces latency by 25.2\% (video) and 43.4\% (community) with no loss in task success rate, and reduces peak memory by 63.0\%. Supported by (E2.2); results in Section~6.3 and \textbf{Figures~12 \& 13}.
    \item \textbf{(C6) Long Context Selection:} {\sys} reduces latency by 12\% vs.\ HF Reranker and 57.3\% vs.\ No Reranker with marginally better precision, and reduces peak memory by $\sim$1\,GiB vs.\ HF Reranker. Supported by (E2.3); results in Section~6.3 and \textbf{Figures~14 \& 15}.
    \item \textbf{(C7) Ablation:} Progressive cluster pruning, chunked execution, dual-layer sliding window, and embedding table caching each achieve their intended optimization goals. Supported by (E3.1); results in Section~6.4 and \textbf{Figure~16}.
\end{itemize}

\subsubsection{Experiments}
Below are the major procedures to reproduce experiments.
\textbf{Please refer to} \texttt{experiments/\\\{experiment\_name\}/README.md} \textbf{for complete instructions}.
\begin{itemize}[leftmargin=*]
    \item \textbf{(E1.1) Latency \& Precision (Figure~8)} [$\sim$10 human-minutes + $\sim$2 compute-hours]. To evaluate the latency and precision of {\sys} vs.\ baselines:
    \begin{itemize}[noitemsep]
        \item \textit{[Execution]}
        {\small\begin{verbatim}
cd experiments/Latency_and_Precision/
bash ./run_latency_experiments.sh
bash ./run_precision_experiments.sh
        \end{verbatim}}
        \item \textit{[Results]}
        {\small\begin{verbatim}
python experiments/Latency_and_Precision/plot.py
        \end{verbatim}}
    \end{itemize}

    \item \textbf{(E1.2) Memory Footprint (Figure~9)} [$\sim$5 human-minutes + $\sim$30 compute-minutes]. To evaluate the peak memory comparison:
    \begin{itemize}[noitemsep]
        \item \textit{[Execution]}
        {\small\begin{verbatim}
bash experiments/Memory_Footprint/run.sh
        \end{verbatim}}
        \item \textit{[Results]}
        {\small\begin{verbatim}
python experiments/Memory_Footprint/plot.py
        \end{verbatim}}
    \end{itemize}

    \item \textbf{(E1.3) Latency--Precision Trade-off (Figure~10)} [$\sim$5 human-minutes + $\sim$1 compute-hour]. To reproduce threshold vs.\ latency/precision curves:
    \begin{itemize}[noitemsep]
        \item \textit{[Execution]}
        {\small\begin{verbatim}
bash ./experiments/Latency_Precision_Tradeoff/run.sh
        \end{verbatim}}
        \item \textit{[Results]}
        {\small\begin{verbatim}
python experiments/Latency_Precision_Tradeoff/plot.py
        \end{verbatim}}
    \end{itemize}

    \item \textbf{(E2.1) RAG Pipeline (Figure~11)} [$\sim$5 human-minutes + $\sim$1 compute-hours]. To evaluate RAG latency and memory:
    \begin{itemize}[noitemsep]
        \item \textit{[Execution]}
        {\small\begin{verbatim}
cd experiments/RAG_Pipeline/
bash ./run_latency_experiments.sh
bash ./run_memory_experiments.sh
        \end{verbatim}}
        \item \textit{[Results]}
        {\small\begin{verbatim}
cd experiments/RAG_Pipeline/
python plot_rag_latency.py
python plot_rag_memory.py
        \end{verbatim}}
    \end{itemize}

    \item \textbf{(E2.2) Agent Memory (Figures~12 \& 13)} [$\sim$5 human-minutes + $\sim$2 compute-hours]. To evaluate Agent Memory latency and memory:
    \begin{itemize}[noitemsep]
        \item \textit{[Execution]}
        {\small\begin{verbatim}
cd experiments/Agent_Memory/
bash ./run_latency_experiments.sh
bash ./run_memory_experiments.sh
        \end{verbatim}}
        \item \textit{[Results]}
        {\small\begin{verbatim}
cd experiments/Agent_Memory/
python plot_agent_latency.py
python plot_agent_memory.py
        \end{verbatim}}
    \end{itemize}

    \item \textbf{(E2.3) Long Context Selection (Figures~14 \& 15)} [$\sim$5 human-minutes + $\sim$1.5 compute-hours]. To evaluate long-context latency and memory:
    \begin{itemize}[noitemsep]
        \item \textit{[Execution]}
        {\small\begin{verbatim}
cd experiments/Long_Context_Selection/
bash ./run_latency_experiments.sh
bash ./run_memory_experiments.sh
        \end{verbatim}}
        \item \textit{[Results]}
        {\small\begin{verbatim}
cd experiments/Long_Context_Selection/
python plot_latency.py
python plot_memory.py
        \end{verbatim}}
    \end{itemize}

    \item \textbf{(E3.1) Ablation (Figure~16)} [$\sim$5 human-minutes + $\sim$1 compute-hours]. To evaluate the ablation study:
    \begin{itemize}[noitemsep]
        \item \textit{[Execution]}
        {\small\begin{verbatim}
cd experiments/Ablation_Study/
bash ./run_latency_ablation.sh
bash ./run_memory_ablation.sh
        \end{verbatim}}
        \item \textit{[Results]}
        {\small\begin{verbatim}
python experiments/Ablation_Study/plot.py
        \end{verbatim}}
    \end{itemize}
\end{itemize}


\end{document}